\documentclass[10pt,twocolumn,letterpaper]{article}

\usepackage{cvpr}              

\usepackage{graphicx}
\usepackage{amsmath}
\usepackage{amssymb}
\usepackage{booktabs}

\usepackage[pagebackref,breaklinks,colorlinks,bookmarks=false,urlcolor=black,citecolor={green!60!black}]{hyperref}

\usepackage{enumitem}
\usepackage[accsupp]{axessibility}

\usepackage[capitalize]{cleveref}
\crefname{section}{Sec.}{Secs.}
\Crefname{section}{Section}{Sections}
\Crefname{table}{Table}{Tables}
\crefname{table}{Tab.}{Tabs.}

\usepackage[dvipsnames]{xcolor}
\usepackage{subcaption}
\usepackage{booktabs}
\usepackage{multirow}
\usepackage{colortbl} 

\usepackage{pifont}  
\usepackage{cite}  
\usepackage{mathtools}  
\usepackage[font=small,labelfont=bf]{caption}  
\usepackage{blindtext}  
\usepackage{bbm}  

\newcommand{\ournet}{CST\xspace}

\newcommand{\textclf}{\text{clf}}
\newcommand{\textseg}{\text{seg}}

\newcommand{\textgt}{\text{gt}}
\newcommand{\textps}{\text{ps}}

\newcommand{\Real}{\mathbb{R}}

\newcommand{\bC}{\mathbf{C}}

\newcommand{\bF}{\mathbf{F}}

\newcommand{\bK}{\mathbf{K}}

\newcommand{\bM}{\mathbf{M}}

\newcommand{\bP}{\mathbf{P}}
\newcommand{\bQ}{\mathbf{Q}}

\newcommand{\bY}{\mathbf{Y}}
\newcommand{\bZ}{\mathbf{Z}}

\newcommand{\bc}{\mathbf{c}}

\newcommand{\bh}{\mathbf{h}}

\newcommand{\bx}{\mathbf{x}}
\newcommand{\by}{\mathbf{y}}

\newcommand{\textq}{{\text{q}}}

\newcommand{\texts}{{\text{s}}}

\newcommand{\gray}[1]{\textcolor{gray}{#1}}

\newcommand{\conditionalcomment}[1]{\if\commenttext1 \else {#1} \fi}
\newcommand{\grayconditionalcomment}[1]{\if\commenttext1 \else \gray{{#1}} \fi}

\newcommand{\lessmarginparagraph}[1]{\smallbreak \noindent \textbf{#1}}

\newcommand{\figref}[1]{Fig.~\ref{#1}\xspace}
\newcommand{\tableref}[1]{Table~\ref{#1}}
\newcommand{\secref}[1]{Sec.~\ref{#1}\xspace}
\renewcommand{\eqref}[1]{Eq.~(\ref{#1})}

\def\etal{\emph{et al.}}
\def\cf{\emph{cf.}\xspace}

\DeclareMathOperator*{\argmax}{arg\,max}

\newcommand{\xmark}{\ding{55}}

\definecolor{grey}{rgb}{0.9, 0.9, 0.9}
\newcommand{\ccol}{\cellcolor{grey}}

\def\cvprPaperID{2019} 
\def\confName{CVPR}
\def\confYear{2023}

\usepackage{txfonts}
\newcommand{\heart}{\ensuremath\varheartsuit}

\begin{document}

\title{Distilling Self-Supervised Vision Transformers \\ for Weakly-Supervised Few-Shot Classification \& Segmentation} 
\author{Dahyun Kang$^{1, 2}$\thanks{Work done during an internship at FAIR.} 
\hspace{3mm}
Piotr Koniusz$^{3, 4}$     \hspace{3mm}
Minsu Cho$^{2}$         \hspace{3mm}
Naila Murray$^{1}$
\\
$^{1}$Meta AI           \hspace{3mm}  
$^{2}$POSTECH           \hspace{3mm}
$^{3}$Data61\heart CSIRO      \hspace{3mm}
$^{4}$Australian National University
\\
}
\maketitle



\begin{abstract}
We address the task of weakly-supervised few-shot image classification and segmentation, by leveraging a Vision Transformer (ViT) pretrained with self-supervision.  
Our proposed method takes token representations from the self-supervised ViT and leverages their correlations, via self-attention, to produce classification and segmentation predictions through separate task heads.
Our model is able to effectively learn to perform classification and segmentation in the absence of pixel-level labels during training, using only image-level labels.
To do this it uses attention maps, created from tokens generated by the self-supervised ViT backbone, as pixel-level pseudo-labels. 
We also explore a practical setup with ``mixed" supervision, where a small number of training images contains ground-truth pixel-level labels and the remaining images have only image-level labels.
For this mixed setup, we propose to improve the pseudo-labels using a pseudo-label enhancer that was trained using the available ground-truth pixel-level labels.
Experiments on Pascal-5$^i$ and COCO-20$^i$ demonstrate significant performance gains in a variety of supervision settings, and in particular when little-to-no pixel-level labels are available.
\end{abstract}

\begin{figure}[t!]
    \centering
    \small
    \includegraphics[width=1.0\linewidth]{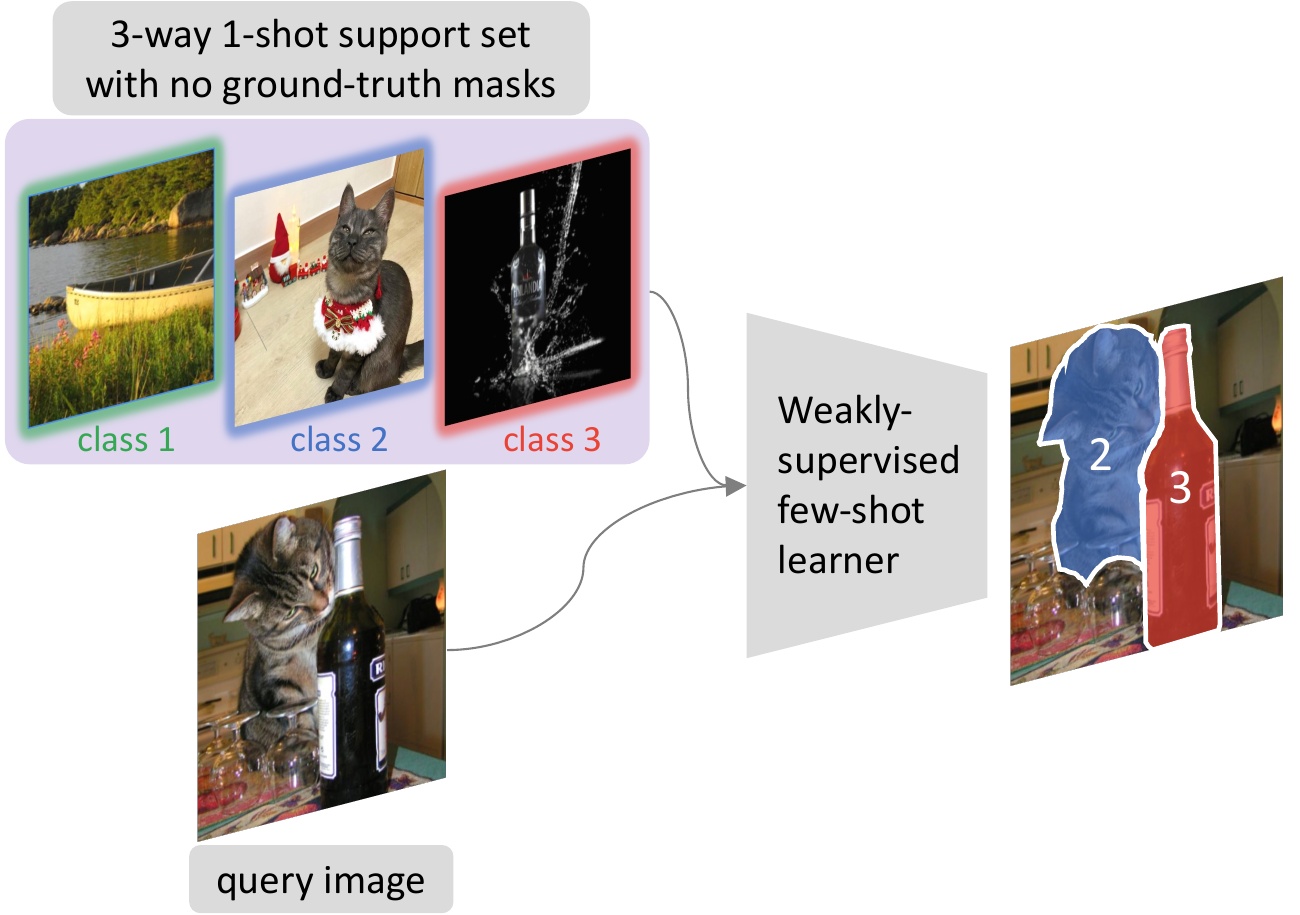}
	\caption{
	\textbf{Weakly-supervised few-shot classification and segmentation.} 
        In this paper, we explore training few-shot models using little-to-no ground-truth segmentation masks.
    }
    \vspace{-6mm}
\label{fig:teaser}
\end{figure}

\section{Introduction}
Few-shot learning~\cite{fink2005object, fei2006one, wang2020generalizing} is the problem of learning to perform a prediction task using only a small number of examples (\ie supports) of the completed task, typically in the range of 1-10 examples.
This problem setting is appealing for applications where large amounts of annotated data are expensive or impractical to collect.
In computer vision, few-shot learning has been actively studied for image classification~\cite{koch2015siamese, matchingnet, relationnet, maml, feat, negmargin} and semantic segmentation~\cite{shaban2017oslsm, dong2018few, zhang2019canet, siam2019amp, zhang2019pgnet}. 
In this work, we focus on the combined task of few-shot classification and segmentation (FS-CS)~\cite{ifsl}, which aims to jointly predict (i) the presence of each support class, \ie, multi-label classification, and (ii) its  pixel-level semantic segmentation.

Few-shot learners are typically trained either by  meta-learning~\cite{schmidhuber1987evolutionary, maml, protonet, wang2019panet} or by transfer learning with fine-tuning~\cite{closer, dhillon2019baseline, chen2021meta}. 
Both these paradigms commonly assume the availability of a large-scale and fully-annotated training dataset. 
For FS-CS, we need Ground-Truth (GT) segmentation masks for query and support images during training.
We also need these GT masks for support images during testing. 
A natural alternative to collecting such expensive segmentation masks  
 would be to instead employ a weaker supervision, \eg image-level~\cite{qi2016augmented, shimoda2016distinct} or box-level labels~\cite{dai2015boxsup, hsu2019weakly}, for learning, \ie, to adopt a weakly-supervised learning approach~\cite{pathak2014fully, li2016weakly, bilen2016weakly, lin2016scribblesup}.   
Compared to conventional weakly-supervised learning however, weakly-supervised {\em few-shot} learning has rarely been explored in the literature and is significantly more challenging.
This is because in few-shot learning, object classes are completely disjoint between training and testing, resulting in models which are susceptible to severe over-fitting to the classes seen during training. 

In this work we tackle this challenging scenario, addressing weakly-supervised FS-CS where only image-level labels, \ie class labels, are available (\cf \figref{fig:teaser}).
Inspired by recent work by Caron \etal~\cite{dino}, which showed that pixel-level semantic attention emerges from self-supervised vision transformers (ViTs)~\cite{vit}, we leverage attention maps from a frozen self-supervised ViT to generate pseudo-GT segmentation masks.
We train an FS-CS learner on top of the frozen ViT using the generated pseudo-GT masks as pixel-level supervision.   
The FS-CS learner is a small transformer network that takes features from the ViT as input and is trained to predict the pseudo-GT masks.
Our complete model thus learns to segment using its own intermediate byproduct, with no segmentation labeling cost, in a form of distillation~\cite{hinton2015distilling,  zhang2020selfdistillation, zhang2021selfdistillation, allen2020towards} between layers within a model.

We also explore a practical training setting 
where a limited number of training images have both image-level and pixel-level annotations, while the remaining images have only image-level labels.
For this {\em mixed-supervised} setting, we propose to train an auxiliary mask enhancer that refines attention maps into pseudo-GT masks. 
The enhancer is supervised using the small number of available GT masks and is used to generate the final pseudo-GT masks for the training images that do not have GT masks.

Lastly, to effectively address this weakly-supervised FS-CS task, we propose a Classification-Segmentation Transformer (\ournet) architecture.
\ournet takes as input ViT tokens for query and support images, computes correlations between them, and then predicts classification and segmentation outputs through separate task heads; the classification head is trained with class labels while the segmentation head is trained with either GT masks or pseudo-GT masks, depending on availability. 
Unlike prior work using ViTs that generate prediction for a single task using either the class token~\cite{vit, pvt} or image tokens~\cite{strudel2021segmenter, pan2021scalable},
\ournet uses each of them for each task prediction, which proves advantageous for both tasks. 
\ournet achieves moderate-to-large performance gains for FS-CS on all three supervision levels (image-level-, mixed-level, and pixel-level), when compared to prior state-of-the-art methods.

Our contributions are the following:
\renewcommand{\labelenumi}{\roman{enumi}.}
\begin{enumerate}[leftmargin=0.6cm]
    \item We introduce a powerful baseline for few-shot classification and segmentation (FS-CS) using only image-level supervision, which leverages localized semantic information encoded in self-supervised ViTs~\cite{dino}.
    \item We propose a learning setup for FS-CS with mixed supervision, and present an auxiliary mask enhancer to improve performance compared to image-level supervision only.
    \item We design Classification-Segmentation Transformer, and a multi-task learning objective of classification and segmentation, which is beneficial for both tasks, and allows for flexibly tuning the degree of supervision.
\end{enumerate}


\vspace{-2mm}
\section{Related work}

\lessmarginparagraph{Few-shot classification and segmentation.}
This paper tackles the recently proposed task of few-shot classification and segmentation (FS-CS)~\cite{ifsl}, which generalizes and combines the conventional few-shot classification (FS-C)~\cite{matchingnet, protonet, maml, closer, renet} and few-shot segmentation (FS-S)~\cite{shaban2017oslsm, wang2019panet, zhang2019canet} tasks.
FS-C assumes that a query contains only \textit{one} of the support classes, while FS-S assumes that the query always contains \textit{all} of them.
In contrast, FS-CS breaks the assumption that a query image contains one class (or all classes) from the support set.
As a result, FS-CS models must predict i) the presence or absence of \textit{each} support class in the query; \textit{and} ii) pixel-level segmentation masks.
Our work explores a more challenging weakly-supervised version of FS-CS with image-level annotations but few or no pixel-level annotations.
Prior work on weakly-supervised FS-S used supervision such as language~\cite{siam2020weakly, lee2022pixel} or bounding boxes~\cite{saha2022improving}.
Our work does not adopt any extra modalities and uses weaker (image-level) supervision than bounding boxes.
Other applications of few-shot learning include few-shot object detection~\cite{zhang2022kernelized, zhang2022time}, action recognition~\cite{wang2022uncertainty, wang2022temporal}, and metric learning~\cite{jung2022few}.

\vspace{-1mm}
\lessmarginparagraph{Weakly-supervised segmentation.}
A common paradigm for weakly-supervised semantic segmentation uses class labels~\cite{lee2021anti, irnet} to generate pseudo-GT masks from class activation maps (CAM)~\cite{classactivationmap, gradcam}, followed by pseudo-supervised training~\cite{zhu2021improving, zou2020pseudoseg, lee2013pseudo}.
CAM requires a classification backbone pre-trained from many examples on \textit{seen} classes whereas, in our few-shot setting, we devise a pseudo-GT mask generation method that requires only a small number of examples, and can handle \textit{unseen} classes.
We also investigate the extent to which a pseudo-mask enhancement module, trained using a small amount of GT masks, can generate better pseudo-GT masks. 
While such ``mixed-supervision" learning has been called semi-weakly-supervised segmentation~\cite{lee2019ficklenet, li2018tell, wang2015semi, bae2022one} and studied in the standard seen-class many-shot learning paradigm, our work is the first to propose mixed-supervision learning for few-shot semantic segmentation, to the best of our knowledge.

\begin{figure*}[t!]
	\centering
	\small
    \includegraphics[width=\linewidth]{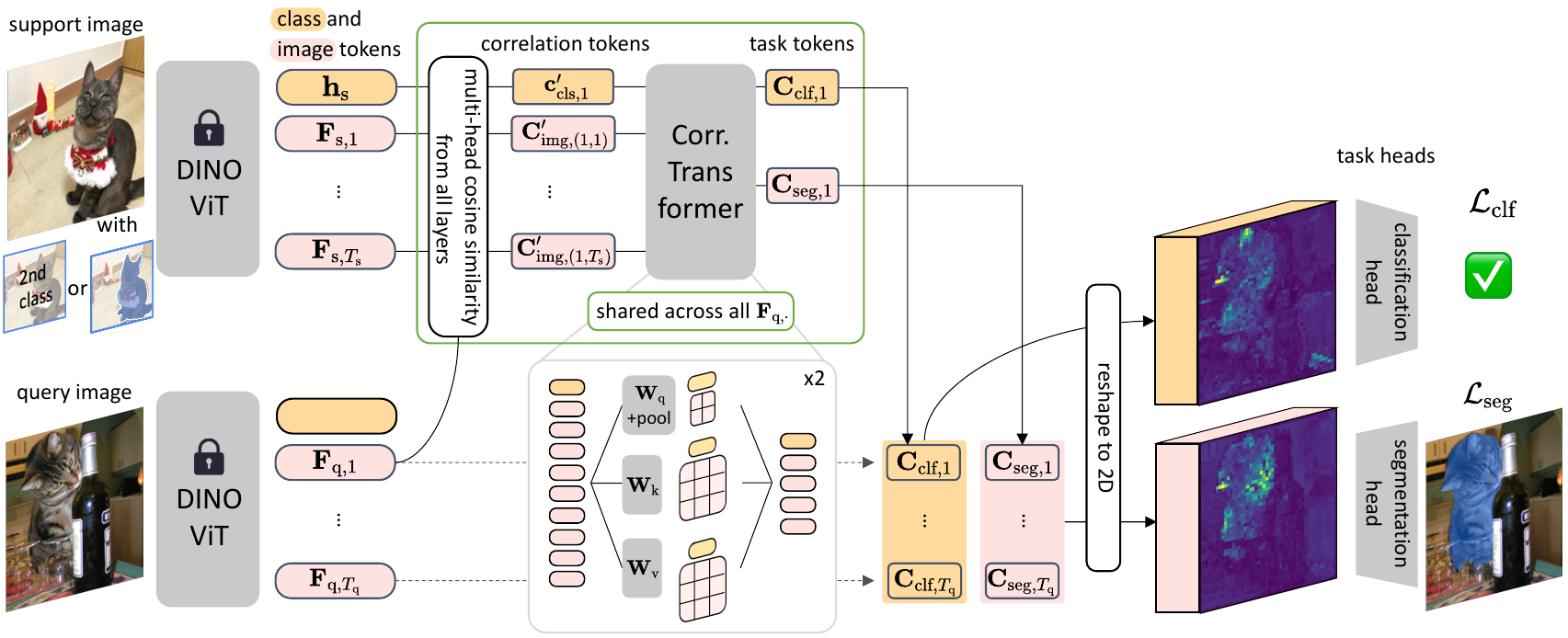}
	\caption{\textbf{Classification-Segmentation Transformer.}
A query and a support image pass through a frozen DINO ViT to extract class and images tokens, which are cross-correlated to form correlation tokens.
The correlation tokens are then fed to a correlation transformer and transform to task tokens.
At the end, each classification and segmentation head takes each task token map and predicts each task. 
}
\label{fig:archi}
\vspace{-4mm}
\end{figure*}

\lessmarginparagraph{Self-supervised learning.}
Self-supervised learning~\cite{jing2020self} has proven highly effective for learning powerful representations for downstream tasks~\cite{wu2018unsupervised, deepcluster, zhuang2019local, chen2020simple, swav,dino}.
One popular approach~\cite{dino, grill2020bootstrap} trains a model to map an image and its augmented views to nearby coordinates in a representation space.
This class-agnostic objective trains the model to capture view-invariant semantics that transcend designated image categories.
In addition, DINO-trained ViTs~\cite{dino} capture localized semantics in the learned token representations.
In this work, we leverage these representations to create attention map-based pseudo-GT masks for weakly-supervised FS-CS, using a novel and efficient self-distillation approach.
The representations have also benefited several other dense prediction tasks, including unsupervised object discovery~\cite{lost, wang2022self} and unsupervised image/video segmentation~\cite{dino, ziegler2022self, zadaianchuk2022unsupervised, yin2022transfgu}.

\section{Weakly-supervised few-shot classification and segmentation}
\label{sec:method}
In this paper we tackle few-shot classification and segmentation (FS-CS)~\cite{ifsl}. 
Given a small number $K$ of labeled support images for each of $N$ classes, an FS-CS model is trained to i) identify the presence of each class in a query image and ii) segment the associated pixels.
This setting is conventionally called an $N$-way $K$-shot episode~\cite{matchingnet, ravi2016optimization}.
We adopt a meta-learning approach~\cite{hochreiter2001learning, matchingnet, protonet} to train our model, where an annotated training dataset (with a class set disjoint from the test set classes) is used to generate a sequence of $N$-way $K$-shot episodes.\footnote{Note that the terms, meta-training and meta-testing, are used interchangeably with training and testing in this paper.}
In our setting, annotations consist of class labels and segmentation masks.
When ground-truth segmentation masks are available for all support and query images during training, we refer to this as pixel-level supervision.
When only class labels are available we refer to this as image-level supervision.
When ground-truth segmentation masks are available for only a small percentage of training images, we refer to this as mixed supervision.

In the remainder of this section, we first introduce our base model architecture, which we call Classification-Segmentation Transformer (\ournet), and then describe training paradigms for the different levels of supervision mentioned above.

\subsection{Classification-Segmentation transformer}
\label{sec:archi}
Our Classification-Segmentation Transformer (\ournet) architecture first computes the cross-correlation of DINO ViT~\cite{dino, vit} tokens between support and query image pairs.
It then generates two new sets of tokens which serve as input to the classification and segmentation prediction heads respectively.
The objective function then combines the losses from each task head.
The details of \ournet's architecture are shown in \figref{fig:archi} and described in the following paragraphs.

\lessmarginparagraph{Correlation tokens.}
We use a frozen ViT architecture pre-trained with DINO as a backbone, which takes as input an image 
and produces $T=HW$ image (patch) tokens $\bF \in \Real^{C\times T}$ and one class token $\bh\in\Real^C$, where $C$ denotes the token feature dimensionality.
The terms ``image" and ``class" tokens are used in analogous fashion to previous work~\cite{bert, vit, deit, dino}. 
The backbone first transforms a query image $q$ and a support image $s$ into two sets of tokens, $\{\bF_\textq, \bh_\textq \}$ and $\{ \bF_\texts, \bh_\texts \}$.
The query image tokens $\bF_\textq$ are cross-correlated, by cosine similarity,
\footnote{The cosine similarity between two vectors is the uncentered Pearson correlation coefficient.}
with the support class token $\bh_\texts$ and the support image tokens $\bF_\texts$, to generate a set of \textit{correlation tokens} $\bc_{\text{cls}}\in\Real^{T_\textq}$ and $\bC_{\text{img}}\in\Real^{T_\textq \times T_\texts}$ respectively:
\begin{equation}
\bc_{\text{cls}} = \overline{\bF}^{\top}_{\textq} \overline{\bh}_\texts, \;\;
\bC_{\text{img}} = \overline{\bF}^{\top}_{\textq} \overline{\bF}_\texts, 
\end{equation}
where $\overline{\bF}$ and $\overline{\bh}$ denote $l2$-normalized tokens.
We retain the semantic diversity of the $M$ heads of the ViT backbone by computing $M$ separate correlation tokens. 
In addition, prior work~\cite{hypercolumns, fpn} has demonstrated the value of incorporating multi-layer semantics. 
Similarly, we compute correlation tokens for each of the $L$ transformer layers of the ViT.
In all, we compute $ML$ different sets of correlation tokens and concatenate them along a new dimension to generate:
\begin{equation}
\bc_{\text{cls}}' \in \Real^{T_\textq \times ML}, \;\; 
\bC_{\text{img}}' \in \Real^{T_\textq \times T_\texts \times ML}. \label{eq:headwisecorr}
\end{equation}

\lessmarginparagraph{Correlation transformer.}
We first concatenate $\bc_{\text{cls},i}' \in \Real^{ML}$ and $\bC_{\text{img},i}' \in \Real^{T_\texts \times ML}$, where $i\in [1,\cdots,T_\textq]$ is an index over the query image tokens, to produce $\bZ_{0,i}\in\Real^{(1+T_\texts) \times ML}$.
Our correlation transformer is a 2-layer transformer~\cite{transformers} that takes $\bZ_{0,i}$ and a segmentation mask label for the support image ($\bM_\textgt \in \{0, 1\}^{T_{\texts}}$) as input, and returns two tokens - a classification task token and a segmentation task token - for query image token $i$.
Each layer has the following form:
\begin{align}
\bZ_{j,i}' &= \mathrm{Norm}^\prime_j(\mathrm{MSA}_j(\bZ_{j,i}, \bM_\textgt) + \bZ_{j,i}), \label{eq:masked_msa}\\
\bZ_{j + 1,i}  &= \mathrm{Norm}_j(\mathrm{MLP}_j(\bZ_{j,i}') + \bZ_{j,i}'),
\end{align}
where $\mathrm{MSA}$, $\mathrm{Norm}$, and $\mathrm{MLP}$ correspond to a multi-head self-attention, a group normalization~\cite{groupnorm}, and a linear layer, respectively, and $j$ denotes the transformer layer index.
Additionally, in each MSA layer, the generated query embeddings $\bQ$ are spatially pooled (using a 4$\times$4 and 3$\times$3 kernel in layers 1 and 2 respectively) by reshaping $\bQ$ to recover its $H_\texts,W_\texts$ dimensions, applying spatial pooling, and flattening the output.
This progressively reduces the number of image tokens, collapsing the input spatial size from $T_\texts=12\times 12$ to 1.
This spatial pooling, illustrated in \figref{fig:archi}, reduces memory consumption, similarly to \cite{dai2020funnel}.
The output of the correlation transformer is then $\bZ_{2,i}\in\Real^{2\times C'}$, where $C'$ denotes the new token dimensionality after MLP projection.
This process is broadcast over the correlation tokens $\bC_{\text{img},i}'$ for each query image token $i$, resulting in $\bZ_2\in\Real^{T_\textq \times 2 \times C'}$.
We then split this tensor, along the second dimension, into classification and segmentation tokens $\bC_\textclf, \bC_\textseg\in\Real^{T_\textq\times C'}$.
Lastly, $\bC_\textclf$ and $\bC_\textseg$ are reshaped into spatial dimensions as $\bC_\textclf, \bC_{\text{seg}} \in \Real^{H_\textq \times W_\textq \times C'}$ and then fed to their respective task heads, described next.

\lessmarginparagraph{Classification and segmentation heads.}
The classification head consists of two 1$\times$1 convolutions followed by global average pooling, and estimates the class presence probability $\by_\textclf$ of the query image corresponding to the support image class.
The segmentation head consists of two 3$\times$3 convolutions followed by bilinear interpolation to the original query image size and predicts a segmentation map $\bY_\textseg$.

\lessmarginparagraph{Learning objective.}
Given predictions from the task heads, the classification and segmentation losses are computed as the cross-entropy between the prediction and the ground-truth for each task:
\begin{align}
    \mathcal{L}_{\textclf} &= -\by_{\textgt} \log \by_\textclf,  \label{eq:loss_cls} \\
    \mathcal{L}_{\textseg} &= - \frac{1}{H W} \sum_{\scriptscriptstyle{\bx \in [H] \! \times \! [W]}} \bY_\textgt(\bx) \log \bY_{\textseg}(\bx),  \label{eq:loss_seg}
\end{align} 
where $\by_\textgt\in\{0,1\}$ is the image-level ground-truth label and $\bY_\textgt(\bx) \in\{0,1\}$ is the ground-truth label for each pixel $\bx$. 
The final learning loss combines the two losses so that the model learns both tasks jointly:
\begin{align}
\mathcal{L} = \lambda \mathcal{L}_\textclf + \mathcal{L}_\textseg,
\label{eq:loss_final}
\end{align}
where $\lambda$ is a balancing hyperparameter.

\lessmarginparagraph{Inference.}
For $N$-way $K$-shot inference, $K$ class logits and segmentation masks are generated using $K$ support images, for each of $N$ classes. These are then averaged, resulting in one estimate per class $n$, $\hat{\by}^{(n)}_\textclf$ and $\hat{\bY}^{(n)}_\textseg$.
Multi-label classification $\tilde{\by}^{(n)}_\textclf$ is then obtained by thresholding each class response $\hat{\by}^{(n)}_\textclf$
with an arbitrarily chosen threshold of $\delta=0.5$.

For segmentation, a pixel $\bx$ is labeled as the class with the highest response, or as background (\ie the $N + 1$-th class) if none of the $N$ class responses exceeds $\delta$. Formally:
\begin{align}
\tilde{\bY}_\textseg(\bx) &= 
\begin{cases}
  N + 1 \text{ \quad if \;}  \forall \; \{ \hat{\bY}^{(n)}_\textseg(\bx)  \}_{n=1}^{N} \leq \delta, \\
  \argmax_{n \in [1, \cdots, N]} \{ \hat{\bY}^{(n)}_\textseg(\bx) \}_{n=1}^{N}. 
\end{cases}
\label{eq:pred_seg}
\end{align}

\subsection{Learning with image-level supervision}
\label{sec:clssup}
\begin{figure}[t!]
	\centering
	\small
    \includegraphics[width=\linewidth]{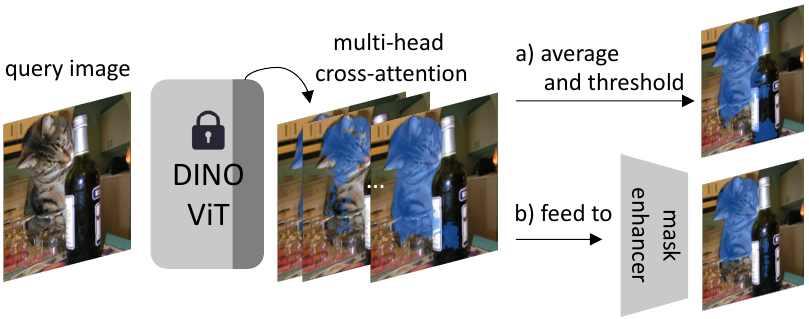}
	\caption{\textbf{a) Image-level-supervision learners} generate pseudo-GT masks by averaging multi-head cross-(self-) attention followed by thresholding (\secref{sec:clssup}).
 \textbf{b) Mixed-supervision learners} first pre-train a mask enhancer with a few GT masks during training to enhance the pseudo-GT mask quality (\secref{sec:mixedsup}). 
}
\label{fig:weaker}
\vspace{-5mm}
\end{figure}
The previously-described learning objective assumes access to GT support and query segmentation mask labels, $\bM_{\textgt}$ and $\bY_{\textgt}$.
To remove the dependence on such expensive annotations, in this section we propose a learning framework for FS-CS that relies only on image-level annotations, \ie, ground-truth class labels.
Using the same model architecture, we adopt a method akin to self-distillation~\cite{zhang2020selfdistillation, zhang2021selfdistillation}, that generates pseudo-GT masks, $\bM_{\textps}$ and $\bY_{\textps}$, and uses them in place of GT mask labels in equations~\ref{eq:masked_msa} and \ref{eq:loss_seg}.

\lessmarginparagraph{Pseudo-GT mask generation.}
As empirically investigated in previous studies~\cite{dino, amir2021deep}, query-key attention maps can capture semantically salient foreground objects.
Inspired by this, we generate pseudo-GT masks for query and support images on the fly by computing the cross- and self-attention from the last DINO ViT layer.
We intercept the key and query embeddings $\bK^m_\texts, \bQ^m_\texts\in\Real^{(1+T_\texts) \times C}$ and $\bK^m_\textq, \bQ^m_\textq\in\Real^{(1+T_\textq) \times C}$ for the support and query images respectively, for the $m$-th head from the last ViT layer.
The support and query image pseudo-GT mask values $\bM_{\textps,i_\texts}$ and $\bY_{\textps,i_\textq}$ (where $i_\texts$ and $i_\textq$ index the support and query image tokens respectively) are then computed as:
\begin{align}
\begin{split}
    \bP_\texts\in\Real^{M \times T_\texts} &; \quad \bP^m_{\texts,i_\texts}={\overline{\bK}^m_{\texts,i_\texts}}^\top \overline{\bQ}^m_{\texts,c_\texts}, \\
    \bP_\textq\in\Real^{M \times T_\textq} &; \quad \bP^m_{\textq,i_\textq}={\overline{\bK}^m_{\textq,i_\textq}}^\top \overline{\bQ}^m_{\texts,c_\texts}, \\ 
    \label{eq:enhancer}
\end{split} 
\end{align} 
\vspace{-8mm}
\begin{align}
\begin{split}
    \bM^\prime_{\textps,i_\texts} &= \frac{1}{M} \sum_{m=1}^{M} \bP^m_{\texts,i_\texts}, \\
    \bY^\prime_{\textps,i_\textq} &= 
    \begin{cases}
    \bf{0} \quad \text{if} \quad \by_\textclf = 0 \\
    \frac{1}{M} \sum_{m=1}^{M} \bP^m_{\textq,i_\textq},
    \end{cases}
\end{split}
\label{eq:pmask}
\end{align}
\begin{equation}
    \bM_{\textps}= \Pi \left( \xi \left(\bM^\prime_{\textps}\right),\alpha \right); \; 
    \bY_{\textps}= \Pi \left( \xi \left(\bY^\prime_{\textps}\right),\alpha \right), \label{eq:pmask_thr}
\end{equation}
with $c_\texts$ denoting the index of the class token for the support image, $\xi(\cdot)$ performing spatial reshaping and bilinear interpolation to the input image size, and $\Pi(\cdot,\alpha)$ applying an element-wise binary threshold $\alpha=-0.1$ (set via hyperparameter search).
The semantics captured in the support class token are matched with the localized semantics of the (query or support) image tokens, using (cross- or self-) attention.
This localized semantic matching is then captured in the resulting pseudo-GT masks.
Our pseudo-mask generation process brings negligible extra computation by leveraging internal byproducts from the ViT backbone.

This learning framework also relates to self-knowledge distillation~\cite{zhang2020selfdistillation, zhang2021selfdistillation}, in the sense that our model is supervised using its own (intermediate) features.
We can think of the backbone as the teacher and the full \ournet model as the student.
However, a key difference is that in our approach the student model learns additional parameters.
In contrast, in strictly self-knowledge distillation approaches the student and teacher model architectures are identical.

\subsection{Extension to mixed-supervision learning}
\label{sec:mixedsup}
As shown in \secref{sec:exp}, our pseudo-masks provide effective supervision for segmentation.
However, there is unsurprisingly still a quality gap between these masks and the GT segmentation masks, which degrades segmentation performance.
In this section, we propose a practical learning setting for FS-CS with \textit{mixed supervision}, which reduces this degradation in segmentation performance, by using a small number of GT segmentation masks during training.
Mixed-supervised FS-CS assumes that all training images contain image-level labels, and a small subset of them also contains pixel-level labels, which are relatively inexpensive compared to full supervision. 

To obtain higher-quality pseudo-GT masks, we use the small number of images with GT masks to train a \textit{mask enhancer}.
We reshape $\bP_\texts$ and $\bP_\textq$ in \eqref{eq:enhancer} into spatial tensors and use them as inputs to the enhancer, which returns binary masks.
During training the enhancer is supervised with the available GT masks. 
The mask enhancer is a network of three 3$\times$3 convolutional layers followed by a Sigmoid.
For mixed-supervision learning, we pre-train this mask enhancer and all pseudo-masks are obtained via the enhancer before being used as supervision 
during meta-training and meta-testing.
Figure~\ref{fig:weaker} compares the two proposed pseudo-GT mask generation processes under the image-level- and mixed-supervision learning settings.

\begin{figure}[t!]
	\centering
	\small
    \includegraphics[width=\linewidth]{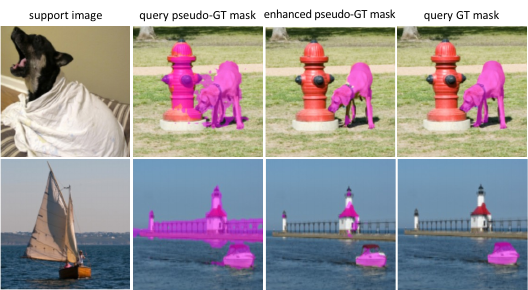}
    \vspace{-6mm}
	\caption{Examples of pseudo-GT \& enhanced pseudo-GT masks. 
}
    \vspace{-3mm}
\label{fig:enhancer}
\end{figure}

\begin{table*}[t!]
    \centering
    \small
    \setlength{\tabcolsep}{5pt}
    \scalebox{0.75}{
        \centering
        \begin{tabular}{lcccccc|ccccc||ccccc|ccccc|r}
            \toprule
             & &\multicolumn{10}{c}{1-way 1-shot} & \multicolumn{10}{c}{2-way 1-shot}\\
             \cmidrule(lr){3-12}\cmidrule(lr){13-22}
             & & \multicolumn{5}{c}{classification 0/1 exact ratio (\%)} & \multicolumn{5}{c}{segmentation mIoU (\%)} & \multicolumn{5}{c}{classification 0/1 exact ratio (\%)} & \multicolumn{5}{c}{segmentation mIoU (\%)} & learn. \\
             \cmidrule(lr){3-7}\cmidrule(lr){8-12}\cmidrule(lr){13-17}\cmidrule(lr){18-22}  method & ps-mask & $5^{0}$ & $5^{1}$ & $5^{2}$ & $5^{3}$ & avg. & $5^{0}$ & $5^{1}$ & $5^{2}$ & $5^{3}$ & avg. & $5^{0}$ & $5^{1}$ & $5^{2}$ & $5^{3}$ & avg. & $5^{0}$ & $5^{1}$ & $5^{2}$ & $5^{3}$ & avg. & params. \\
             \midrule
             HSNet~\cite{hsnet} & \xmark & \textbf{84.5} & \textbf{84.8} & 60.8 & \textbf{85.3} & 78.9 & 20.0 & 23.5 & 16.2 & 16.6 & 19.1 & 70.8 & 67.0 & 36.5 & \textbf{70.1} & 61.1 & 11.0 & 23.4 & 15.4 & 17.0 & 17.7 & 2.6M \\
             ASNet~\cite{ifsl} & \xmark & 80.2 & 84.0 & 66.2 & 82.7 & 78.3 & 11.7 & 21.1 & 13.4 & 16.2 & 15.6 & 67.6 & \textbf{70.2} & 44.5 & 69.3 & 62.9 & 10.3 & 20.7 & 12.6 & 15.9 & 14.9 & 1.3M  \\
             \ournet (image) & \xmark & 83.9 & 80.7 & 65.9 & 82.7 & 78.3 & 11.9 & 20.7 & 15.0 & 16.3 & 16.0 & 72.6 & 65.3 & 43.2 & 69.5 & 62.6 & 11.9 & 20.5 & 15.0 & 16.6 & 15.7 & 0.4M \\
             DINO~\cite{dino} & $\diamond$ & - & - & - & - & - & 20.0 & 23.4 & 16.2 & 16.6 & 19.1 & - & - & - & - & - & 12.8 & 12.9 & 10.0 & 9.9 & 11.4 & 0 \\
             \ournet & \checkmark & 84.0 & 82.2 & \textbf{70.8} & 82.6 & \textbf{79.9} & \textbf{35.8} & \textbf{38.9} & \textbf{28.9} & \textbf{29.2} & \textbf{33.2} & \textbf{74.3} & 67.2 & \textbf{49.0} & 67.8 & \textbf{64.6} & \textbf{35.7} & \textbf{36.0} & \textbf{26.8} & \textbf{29.1 }& \textbf{31.9} & 0.4M \\
             \bottomrule
        \end{tabular}
    }
    \vspace{-2mm}
    \caption{Comparing model performance on FS-CS when trained with image-level supervision on Pascal-5$^i$~\cite{shaban2017oslsm}. ``ps-mask'' refers to using pseudo-GT masks during training. Our \ournet model compares favourably to prior works on classification, and outperforms them on segmentation. Results for DINO~\cite{dino} were generated without any training on Pascal-5$^i$ (denoted by $\diamond$). See section~\ref{sec:exp_ils} for details.
}
    \label{table:p_cls}
    \vspace{-2mm}
\end{table*}

\begin{table}[t!]
    \centering
    \small
    \setlength{\tabcolsep}{4pt}
    \scalebox{0.92}{
        \begin{tabular}{lccc||cccc}
            \toprule
             & & \multicolumn{2}{c}{1-way 1-shot} & \multicolumn{2}{c}{2-way 1-shot}\\
             \cmidrule(lr){3-4}\cmidrule(lr){5-6}
              method & ps-mask & cls. ER & seg. mIoU & clf. ER & seg. mIoU \\
             \midrule
             \ournet  & \xmark &  74.5 & 10.3  & 62.3 & 9.8 \\
             DINO \cite{dino}   & $\diamond$ & - & 12.1 & -  & 7.4 \\
             \ournet  & \checkmark & \textbf{78.2} & \textbf{19.6} & \textbf{62.4} & \textbf{18.3} \\
             \bottomrule
        \end{tabular}
    }
    \vspace{-0mm}
    \caption{
    Comparing model performance on FS-CS when trained with image-level supervision on COCO-20$^i$~\cite{nguyen2019fwb}, averaged over 4 folds. See section~\ref{sec:exp_ils} for details.
    }
    \label{table:c_cls}
    \vspace{-5mm}
\end{table}

\section{Experiments}
\label{sec:exp}
In this section we provide implementation details for our approach, and present quantitative and qualitative results to experimentally evaluate its performance.
The attached supplementary material contains more analyses and details.

\lessmarginparagraph{Architecture and training.}
To compare with prior work based on ResNet50~\cite{wang2019panet, tian2020pfenet, hsnet, ifsl}, we use a ViT-small backbone~\cite{vit} with 21M parameters that was pre-trained via self-supervision with DINO~\cite{dino} on ImageNet 1K~\cite{russakovsky2015imagenet}.
The size and training data are comparable with ResNet50~\cite{resnet}, which has 23M parameters and was also trained on ImageNet 1K, but using class labels as supervision.
ViT-small consists of $L=12$ transformer layers of $M=6$ attention heads, with $C=64$-dimensional query-key-value embeddings.
The backbone is frozen, as in previous work~\cite{tian2020pfenet, zhang2021few}, and the additional layers are trained using Adam~\cite{adam} with a learning rate of $10^{-3}$.
Consistent with~\cite{ifsl}, \ournet is trained with 1-way 1-shot episodes and is used for any $N$-way $K$-shot inference.
We set the task token dimensionality $C'=128$, and the loss balancing hyperparameter $\lambda = 0.1$.

\lessmarginparagraph{Datasets.}
We report results on two standard FS-CS benchmarks~\cite{ifsl}, Pascal-5$^{i}$~\cite{shaban2017oslsm, pascal} and COCO-20$^{i}$~\cite{nguyen2019fwb, coco}.
Both datasets use 4-fold validation, where the folds $i\in\{0,1,2,3\}$ are created by splitting the 20 (resp. 80) classes into 4 mutually disjoint class sets, with 10K (resp. 100K) images per fold.
For training and validation, support and query image pairs are sampled on the fly, as done in \cite{ifsl}. 
For each training episode, $N$ support classes are chosen randomly from the split and $K$ images are chosen randomly from each of the $N$ target classes to construct an $N$-way $K$-shot support set.
When a support image contains multiple classes, only one of them is set as its image class and the rest are set as background.
While this introduces noise into support sets containing multi-class images, we find that it performs well in practice.
A query image is also chosen randomly and may contain between 0 to $N$ of the target classes.
The image-level label for the query is then constructed as an $N$-dimensional binary vector containing 1 if the query contains the $n$-th class, and 0 otherwise.
Similarly, its pixel-level label is defined as an $(N\!+\!1)$-way segmentation mask, including background.
For mixed-supervision learning, we sort the images in a training split by their image names and use pixel-level labels for the first \eg 5\% of images, and only image-level labels for the remainder.
This effectively randomizes the subset of images with full-supervision.

\lessmarginparagraph{Evaluation and metrics.}
We follow the established evaluation protocol for each FS-CS benchmark~\cite{shaban2017oslsm, nguyen2019fwb, ifsl}.
To evaluate multi-label classification, we use the 0/1 exact match ratio~\cite{durand2019learning}, where an image is considered to be correctly classified if the binary label for each of the $N$ classes is correct.
For semantic segmentation, we use mean IoU, $\mathrm{mIoU} = \frac{1}{R}\sum_R \mathrm{IoU}_r$~\cite{shaban2017oslsm, wang2019panet}, where $\mathrm{IoU}_r$ is the IoU of all images labeled with the $r_{\text{th}}$-class.
The best model is chosen based on the highest validation $\mathrm{mIoU}$ averaged with 1000 random validation episodes.

\subsection{Results with image-level supervision}
\label{sec:exp_ils}
Tables~\ref{table:p_cls} and~\ref{table:c_cls} compare the performance of our image-level-supervised \ournet with baseline methods.
The only prior work that addresses this setting~\cite{ifsl} generates a segmentation mask that is then globally average-pooled. 
The cross-entropy between the average-pooled prediction and the ground-truth image label is used as the objective function.
We adopt this approach for~\cite{hsnet}, which did not address this setting, to construct another baseline, and we also evaluate \ournet in this setting, which we denote with \xmark.
The result for DINO is computed by evaluating segmentation masks that were generated as in~\cite{dino} (denoted by $\diamond$), without using any training data from the target datasets.
Our \ournet model, trained with pseudo-masks via self-distillation (denoted by \checkmark), performs comparably to prior work on classification and, because of the superior supervision provided by our pseudo-masks during training, outperforms these models on segmentation.
In particular, \ournet achieves an over 100\% improvement over ASNet~\cite{ifsl} in both the 1-way 1-shot and 2-way 1-shot settings.
We also note that training CST with pseudo-GT masks (\checkmark) also improves classification performance by 1.6\% (1-way 1-shot) and 2\% (2-way 1-shot) on Pascal-5$^{i}$ when compared to training without (\xmark).

\begin{table}[t!]
    \centering
    \small
    \scalebox{0.90}{
        \begin{tabular}{lccccc}
            \toprule
             \% GT training set masks $\rightarrow$ &  0\% & 2.5\% & 5\% & 10\% & 100\% \\ \midrule
             (a): pseudo-mask baseline & 33.2 & 41.4 & 43.5 & 44.5 & - \\
             (b): (a) + mask enhancer & - & 43.2 & 44.1 & 45.2 & - \\
             (c): (b) + GT test set $\mathcal{S}$ masks  & - & 47.5 & 48.4 & 50.1 & 55.5 \\
             \bottomrule
        \end{tabular}
    }
    \vspace{-3mm}
    \caption{Comparing segmentation mIoU of \ournet on FS-CS with mixed supervision on Pascal-5$^i$, for different \% of training set ground-truth masks, averaged over 4 folds. In (c), ground-truth masks were used for all support ($\mathcal{S}$) images during testing.
    }
    \label{table:p_mixed}
    \vspace{-4mm}
\end{table}

\begin{table*}[t!]
    \centering
    \small
    \setlength{\tabcolsep}{5pt}
    \scalebox{0.80}{
        \begin{tabular}{lccccc|ccccc||ccccc|ccccc|r}
            \toprule
             & \multicolumn{10}{c}{1-way 1-shot} & \multicolumn{10}{c}{2-way 1-shot} \\
             \cmidrule(lr){2-11}\cmidrule(lr){12-21} 
             & \multicolumn{5}{c}{classification 0/1 exact ratio (\%)} & \multicolumn{5}{c}{segmentation mIoU (\%)} & \multicolumn{5}{c}{classification 0/1 exact ratio (\%)} & \multicolumn{5}{c}{segmentation mIoU (\%)} & learn. \\
             \cmidrule(lr){2-6}\cmidrule(lr){7-11}\cmidrule(lr){12-16}\cmidrule(lr){17-21} method & $5^{0}$ & $5^{1}$ & $5^{2}$ & $5^{3}$ & avg. & $5^{0}$ & $5^{1}$ & $5^{2}$ & $5^{3}$ & avg. & $5^{0}$ & $5^{1}$ & $5^{2}$ & $5^{3}$ & avg. & $5^{0}$ & $5^{1}$ & $5^{2}$ & $5^{3}$ & avg. & params. \\
             \midrule
             PANet \cite{wang2019panet} & 69.9 & 67.7 & 68.8 & 69.4 & 69.0 & 32.8 & 45.8 & 31.0 & 35.1 & 36.2 & 56.2 & 47.5 & 44.6 & 55.4 & 50.9 & 33.3 & 46.0 & 31.2 & 38.4 & 37.2 & 23.6M \\
             PFENet \cite{tian2020pfenet} & 69.8 & 82.4 & 68.1 & 77.9 & 74.6 & 38.3 & 54.7 & 35.1 & 43.8 & 43.0 & 22.5 & 61.7 & 40.3 & 39.5 & 41.0 & 31.1 & 47.3 & 30.8 & 32.2 & 35.3 & 31.5M \\
             HSNet \cite{hsnet} & 86.6 & 84.8 & 76.9 & 86.3 & 83.7 & 49.1 & 59.7 & 41.0 & 49.0 & 49.7 & 68.0 & 73.2 & 57.0 & 70.9 & 67.3 & 42.4 & 53.7 & 34.0 & 43.9 & 43.5 & 2.6M \\
             ASNet \cite{ifsl} & 84.9 & \textbf{89.6} & 79.0 & 86.2 & 84.9 & 51.7 & 61.5 & 43.3 & 52.8 & 52.3 & 68.5 & 76.2 & 58.6 & 70.0 & 68.3 & 48.5 & 58.3 & 36.3 & 48.3 & 47.8 & 1.3M \\
             \ournet & \textbf{86.9} & 88.0 & \textbf{81.5} & \textbf{86.5} & \textbf{85.7} & \textbf{55.6} & \textbf{61.6} & \textbf{47.7} & \textbf{56.9} & \textbf{55.5} & \textbf{70.3} & \textbf{74.9} & \textbf{62.0} & \textbf{74.4} & \textbf{70.4} & \textbf{56.9} & \textbf{61.0} & \textbf{48.0} & \textbf{57.0} & \textbf{55.7} & 0.4M \\
             \bottomrule
        \end{tabular}
    }
    \vspace{-2mm}
    \caption{Comparing FS-CS models trained with pixel-level supervision on Pascal-5$^{i}$.
}
    \label{table:p_mask}
    \vspace{-4mm}
\end{table*}

\begin{table}[t!]
    \centering
    \small
    \scalebox{0.9}{
        \begin{tabular}{lcc||cc}
            \toprule
             & \multicolumn{2}{c}{1-way 1-shot} & \multicolumn{2}{c}{2-way 1-shot}\\
             \cmidrule(lr){2-3}\cmidrule(lr){4-5}
             method  & clf. ER & seg. mIoU & clf. ER & seg. mIoU \\
             \midrule
             PANet \cite{wang2019panet}   & 66.7 & 25.2 & 48.5 & 23.6 \\ 
             PFENet \cite{tian2020pfenet} & 71.4 & 31.9 & 36.5 & 22.6 \\ 
             HSNet \cite{hsnet}           & 77.0 & 34.3 & 62.5 & 29.5 \\ 
             ASNet \cite{ifsl}            & 78.6 & 35.8 & 63.1 & 31.6 \\
             \ournet & \textbf{80.8} & \textbf{38.3} & \textbf{64.0} & \textbf{36.2} \\
             \bottomrule
        \end{tabular}
    }
    \vspace{-0mm}
    \caption{
    Comparing performance on FS-CS with pixel-level supervision on COCO-20$^i$. 
    }
    \label{table:c_mask}
    \vspace{-2mm}
\end{table}

\subsection{Results with mixed supervision}
To validate the effectiveness of the mixed-supervision setting, we vary the percentage of pixel-level annotated images used in training and present the results in Table~\ref{table:p_mixed}.
We observe that using just 5\% of GT masks is remarkably effective at improving performance, when compared to using 0\% (43.5\% vs. 33.2\%).
The mask enhancer brings much improvement with fewer ground-truth segmentation masks. 
The effect of the mask enhancer is visually demonstrated in \figref{fig:enhancer}.
We observe that the mask enhancer reduces foreground noise and often filters out objects irrelevant to a support image.
Note that mixed-supervision learning assumes access to pixel-level annotations only during training.
During testing, support images use pseudo-masks without (Table~\ref{table:p_mixed} (a)) or with (Table~\ref{table:p_mixed} (b)) mask enhancement.
In (c), we use one-shot support GT mask during testing, but still use reduced amounts of GT masks in training.
In this setting, the performance is only 5.4\%-8.0\% lower than the fully-supervised model, indicating that our approach allows for significant pixel-level annotation efficiency during training.

\begin{figure}[t!]
	\centering
	\small
    \includegraphics[width=\linewidth]{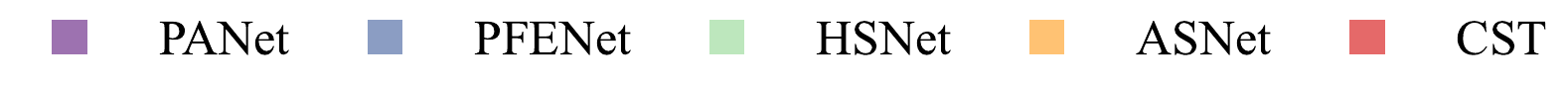}
    \vspace{-2mm}
    \includegraphics[width=0.49\linewidth]{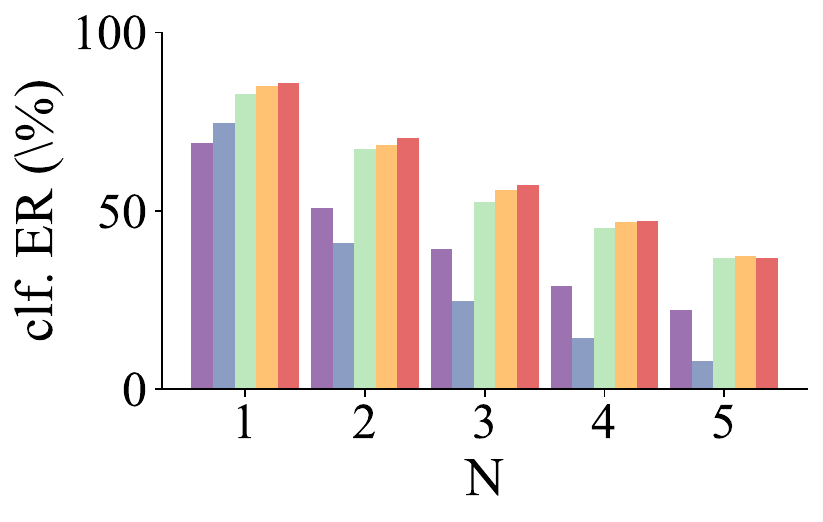} \hfill
    \includegraphics[width=0.49\linewidth]{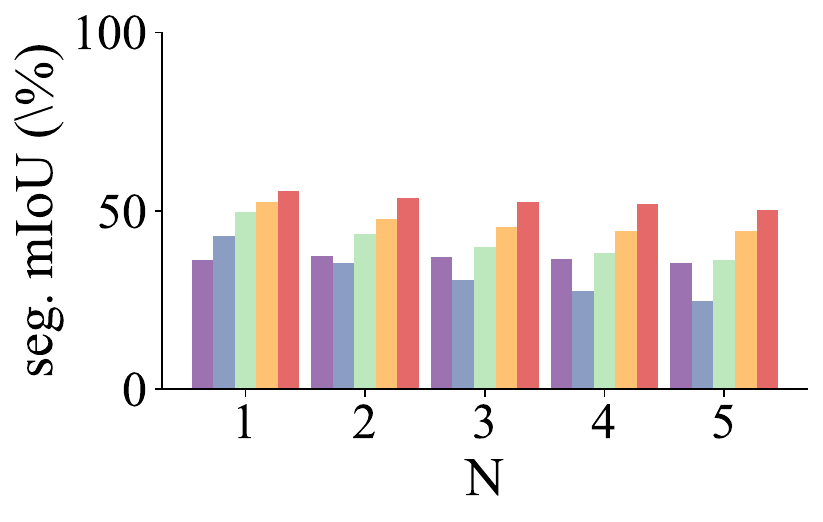}
    \vspace{-2mm}
    \caption{Comparing performance on FS-CS with pixel-level supervision on $N$-way 1-shot Pascal-5$^{i}$.}
    \vspace{-5mm}
\label{fig:p-nway}
\end{figure}

\begin{figure*}[t!]
	\centering
	\small
    \includegraphics[width=0.97\linewidth]{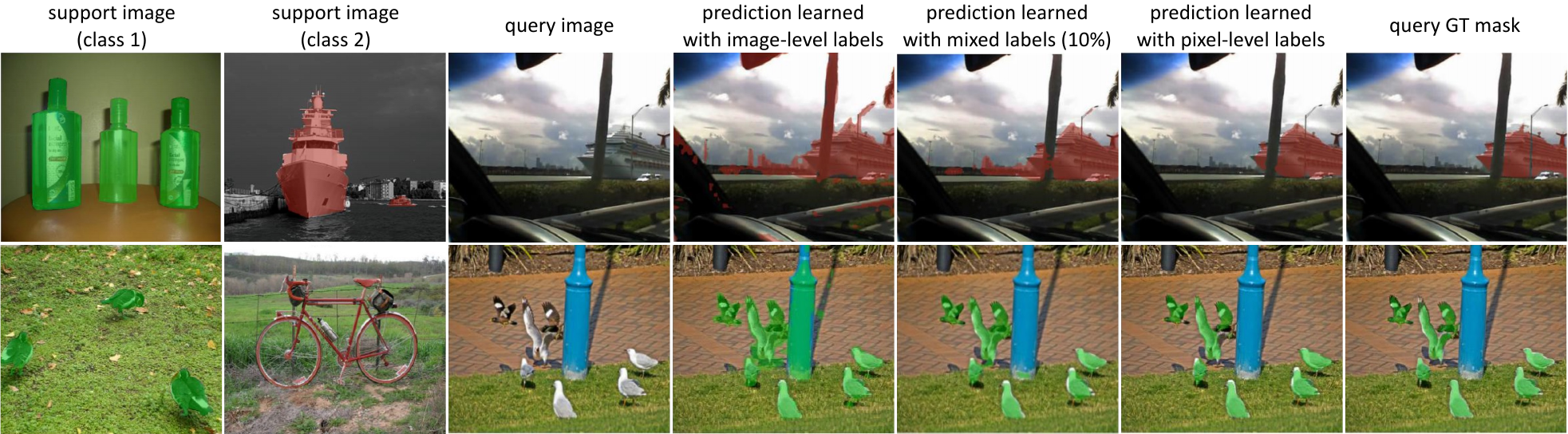}
    \vspace{-2mm}
    \caption{2-way 1-shot segmentation examples from \ournet trained with image-level, mixed, and pixel-level supervision. 
    Note that image-level-supervised and mixed-supervised models do not have access to ground-truth support masks. In both cases, the model is still effective.
    }
    \vspace{-4mm}
\label{fig:main}
\end{figure*}

\subsection{Results with full, \ie pixel-level, supervision}
Tables~\ref{table:p_mask} and \ref{table:c_mask}, as well as \figref{fig:p-nway}, compare our model's performance on FS-CS with state-of-the-art methods whose results, included here, were initially reported in \cite{ifsl}.
\ournet achieves significant improvements over previous work, which we attribute to 
(i) our correlation tokens that leverage the localized semantics encoded in our self-supervised ViT backbone model; and 
(ii) the two task heads of our transformer (see ablations in \secref{sec:exp_abl}).
Figure~\ref{fig:main} visually demonstrates qualitative examples of final segmentation predictions with increasing levels of supervision, from image-level to pixel-level.
Image-level supervised predictions roughly segment out objects that correspond with the support classes, and the predictions become less noisy as more pixel-level supervision is used.

\begin{figure}[t!]
	\centering
	\small
    \includegraphics[width=0.9\linewidth]{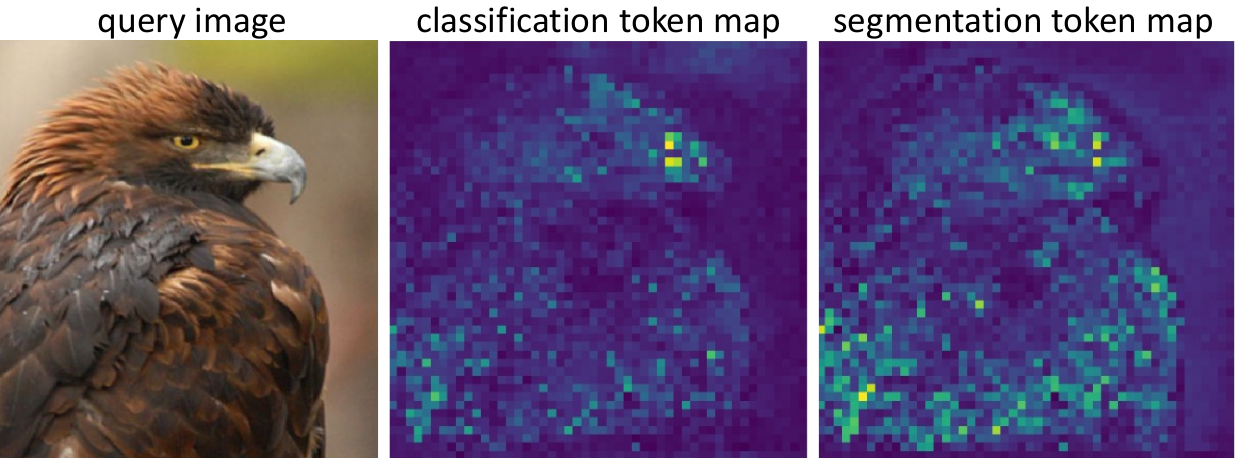}
    \vspace{-2mm}
	\caption{An example of channel-averaged $\bC_\textclf$, $\bC_\textseg$. 
 Each task token tends to capture different aspects, \eg, clearer edges of $\bC_\textseg$.
}
\vspace{-2mm}
\label{fig:twoheads}
\end{figure}

\subsection{Ablation study}
\label{sec:exp_abl}

\begin{table}[t!]
    \centering
    \small
    \setlength{\tabcolsep}{5pt}\
    \scalebox{0.95}{
        \begin{tabular}{lcccc}
            \toprule
             labels $\rightarrow$ & class-sup. & self-sup. & self-sup. & self-sup. \\ 
             archi. $\rightarrow$ & ViT\cite{vit} & CNN\cite{resnet} & ViT\cite{dino}$^{*}$ & ViT\cite{dino} \\ \midrule
             seg. mIoU (\%) & 6.9 & 20.7 & 15.6 & 33.2 \\
             \bottomrule
        \end{tabular}
    }
    \vspace{-0mm}
    \caption{
    FS-CS model comparison trained with image-level labels on 1-way 1-shot Pascal-5$^{i}$ by varying pseudo-mask generation models.
    Different columns denote model architectures for pseudo-labeling and their pre-training supervision levels.
    The method $*$ uses the final layer tokens from \cite{dino} for generating pseudo-labels.
    }
    \label{table:p_mask_abl}
    \vspace{-0mm}
\end{table}

\begin{table}[t!]
    \centering
    \small
    \scalebox{0.98}{
        \begin{tabular}{llcc||cc}
            \toprule
             & & \multicolumn{2}{c}{1-way 1-shot} & \multicolumn{2}{c}{2-way 1-shot}\\
             \cmidrule(lr){3-4}\cmidrule(lr){5-6}
             clf. \& seg.  & superv. & clf. & seg. & clf. & seg. \\
             prediction heads & level & ER & mIoU & ER & mIoU \\
             \midrule
             coupled & image & 78.5 & 32.9 & 59.5 & \textbf{31.9} \\
             \ccol decoupled (\ournet) & \ccol image & \ccol \textbf{79.9} & \ccol \textbf{33.2} & \ccol \textbf{64.5} & \ccol \textbf{31.9} \\
             coupled & pixel & 84.3 & 54.2 & 67.0 & 53.0 \\
             \ccol decoupled (\ournet) & \ccol pixel & \ccol \textbf{85.7} & \ccol \textbf{55.5} & \ccol \textbf{70.5} & \ccol \textbf{55.7} \\
             \bottomrule
        \end{tabular}
    }
    \vspace{-0mm}
    \caption{
    Ablation study of coupling or decoupling classification and segmentation prediction heads for FS-CS on Pascal-5$^{i}$.
    }
    \label{table:p_heads}
    \vspace{-5mm}
\end{table}

\lessmarginparagraph{Pseudo-labeling with different models.}
We explore other learning-free pseudo-labeling methods by using alternative pre-trained feature extractors to generate pseudo-GT masks.
Rather than self-distillation, where we train \ournet with pseudo-GT masks generated from its own self-supervised DINO ViT backbone, in these experiments we train \ournet with pseudo-GT masks generated from class-supervised ViT~\cite{vit}, self-supervised DINO ResNet50~\cite{resnet}, or the output layer tokens from the self-supervised DINO ViT~\cite{dino}.
As ResNets do not have class tokens, we apply global-average pooling on the support feature map from the last ResNet layer and use the pooled feature as a class token for generating the pseudo-GT masks. 
As seen in Table~\ref{table:p_mask_abl}, pseudo-GT masks generated from alternative feature extractors do not perform as well as the intermediate attention maps from DINO.
We qualitatively observed that the alternatives' less accurate pseudo-masks lead to lower segmentation accuracy when used in pseudo-supervised learning:
the self-supervised ResNet50 produces pseudo-masks with indistinct object boundaries, 
while the class-supervised ViT produces noisy patches throughout the pseudo-mask (as was similarly observed in \cite{dino}).

\begin{table}[t!]
    \centering
    \small
    \scalebox{0.86}{
        \begin{tabular}{lcc||ccr}
            \toprule
             & \multicolumn{2}{c}{1-way 1-shot} & \multicolumn{2}{c}{1-way 5-shot} & \small{\# learn.} \\
             \cmidrule(lr){2-3}\cmidrule(lr){4-5}
             method & mIoU & FBIoU & mIoU & FBIoU & \small{params.}\\
             \midrule
             RPMM \cite{yang2020pmm}      & 30.6 & - & 35.5 & - & 38.6M \\ 
             RePRI \cite{malik2021repri}  & 34.0 & - & 42.1 & - & - \\ 
             SSP \cite{fan2022self} & 37.4 & - & 44.1 & - & 8.7M \\ 
             MMNet \cite{wu2021learning}  & 37.5 & - & 38.2 & - & 10.4M \\ 
             MLC \cite{yang2021mining}    & 33.9 & - & 40.6 & - & 8.7M \\ 
             NTRENet \cite{liu2022learning}&39.3 & 68.5 & 40.3 & 69.2 & -\\ 
             CMN \cite{xie2021few}        & 39.3 & 61.7 & 43.1 & 63.3 & - \\ 
             HSNet \cite{hsnet}           & 39.2 & 68.2 & 46.9 & 70.7 & 2.6M \\   
             DACM \cite{xiong2022doubly} & 40.6 & 68.9 & 48.1 & 71.6 & - \\ 
             ASNet \cite{ifsl}                  & 42.2 & 68.8 & 47.9 & 71.6 & 1.3M \\      
             \ournet                  & \textbf{44.0} & \textbf{70.3} & \textbf{48.7} & \textbf{73.7} & 0.4M \\      
             \bottomrule
        \end{tabular}
    }
    \vspace{-2mm}
    \caption{Comparing FS-S models trained with pixel-level supervision on COCO-20$^i$. 
    }
    \label{table:c_fss}
    \vspace{-5mm}
\end{table}

\lessmarginparagraph{Effect of decoupled task heads.}
Table~\ref{table:p_heads} compares the impact of coupling and decoupling the classification and segmentation task heads. 
The coupled architecture has a single segmentation task head, and the image is classified with the foreground class if any pixels are predicted as foreground, and is otherwise classified as the background class.
Having separate heads for each task benefits both tasks in both the image-level and pixel-level supervision settings.
A similar observation was made for object detection~\cite{song2020revisiting} when decoupling localization from classification.
Figure~\ref{fig:twoheads} visualizes an example of channel-averaged $\bC_\text{clf}$ and $\bC_\text{seg}$ tensors.
We observe that these two intermediate representations tend to highlight different areas; $\bC_\text{clf}$ has large responses in highly-localized class-specific image regions, while $\bC_\text{seg}$ better distinguishes the overall object support and boundaries.

\lessmarginparagraph{Comparison on FS-S.}
In \tableref{table:c_fss}, \ournet is also evaluated in the conventional FS-S setting~\cite{shaban2017oslsm, wang2019panet}, \ie, binary segmentation using mask supervision, and compared with previous works that use different ResNet50-based architectures.
FS-S is equivalent to FS-CS with the precondition that a query and a support always share a class label;
classification is not required.
We train \ournet on this setting and find that it achieves state-of-the-art segmentation performance.
It is also worth noting that it achieves this with fewer learned parameters than others, which is also the case for the comparisons in \tableref{table:p_cls} and \tableref{table:p_mask}.
Our small number of learned parameters may help to avoid overfitting.

\section{Conclusion}
In this work, we propose an approach to generating effective pseudo-GT masks for few-shot classification and segmentation (FS-CS) using the localized semantics encoded in the tokens of self-supervised vision transformers.
We show that using these pseudo-GT masks, sometimes combined with a small number of GT masks, we can train effective FS-CS models.
Our work is the first to explore this practical weakly-supervised FS-CS setting, which drastically reduces the amount of expensive pixel-level annotations needed for training and deployment.
Extensive experiments show that our approach achieves significant performance gains when compared to state-of-the-art methods in a variety of supervision settings, on two FS-CS benchmarks.

\lessmarginparagraph{Acknowledgements.}
This work was partly supported by the IITP grants (2022-0-00959: Few-Shot Learning
of Causal Inference in Vision and Language, 2022-0-00113: Sustainable Collaborative Multi-modal Lifelong Learning) funded by Ministry of Science and ICT, Korea.  
PK was in part funded by CSIRO's Machine Learning and Artificial Intelligence Future Science Platform (MLAI FSP) Spatiotemporal Activity.



{\small
\bibliographystyle{ieee_fullname}
\bibliography{dahyuns}
}

\clearpage
\renewcommand{\theequation}{a.\arabic{equation}}
\renewcommand{\thetable}{a.\arabic{table}}
\renewcommand{\thefigure}{a.\arabic{figure}}
\renewcommand*{\thefootnote}{\arabic{footnote}}
\renewcommand\thesection{\Alph{section}}
\setcounter{section}{0}
\setcounter{figure}{0}
\setcounter{table}{0}

\section{Appendix}
We provide additional details and analyses of the proposed method in this supplementary material.

\subsection{Implementation details}
\begin{figure}[t!]
	\centering
	\small
    \includegraphics[width=\linewidth]{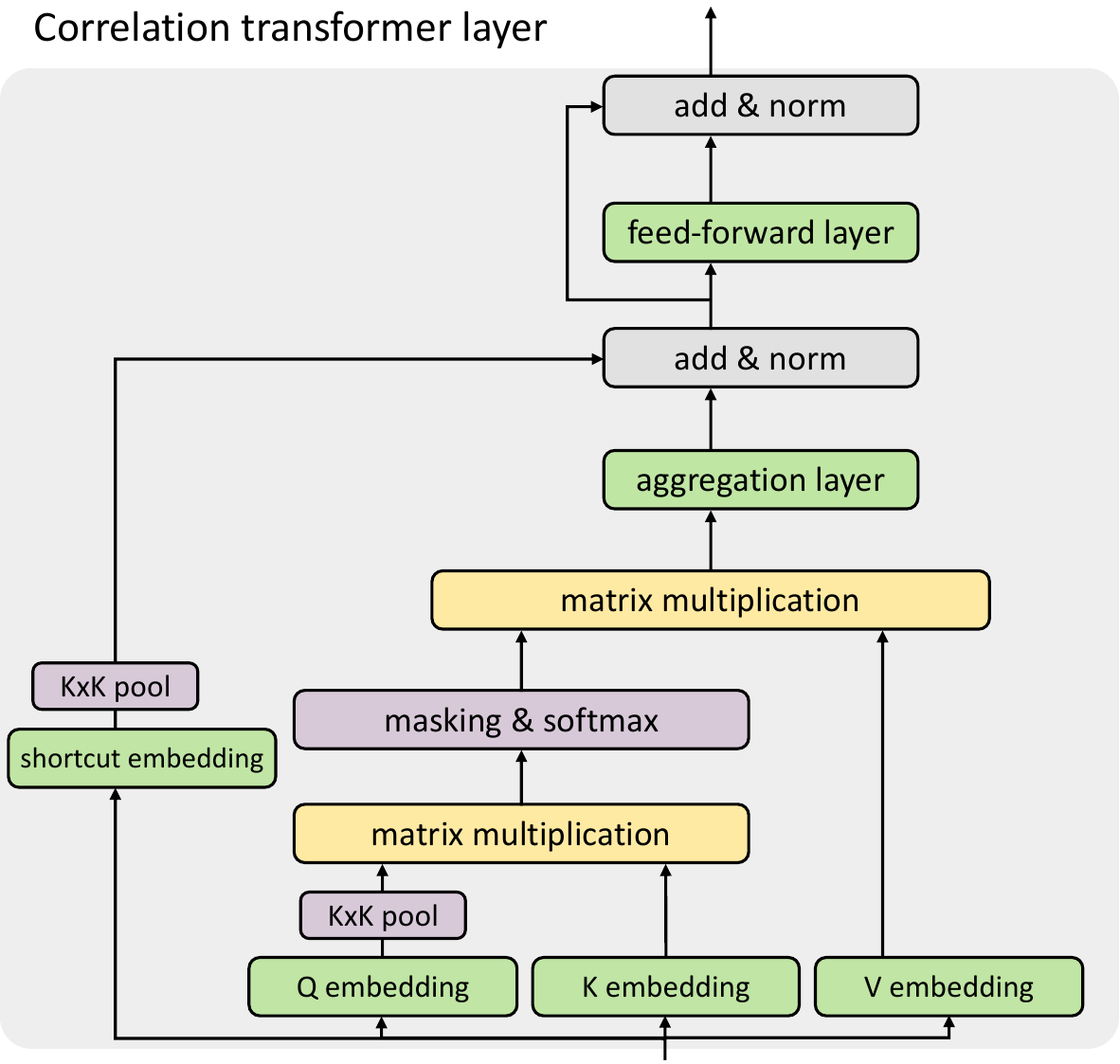}
    \vspace{-5mm}
	\caption{
 Illustration of correlation transformer layer.
 Each query-key-value, shortcut, aggregation, and feed-forward layer is implemented with an MLP layer.
 We use group normalization~\cite{groupnorm} with 4 groups and ReLU~\cite{relu} activation in implementation but omit them in this illustration for simplicity.
    }
\label{fig:corrtrlayer}
\vspace{-3mm}
\end{figure}

Our framework is implemented using the PyTorch Lightning library~\cite{falcon2019pytorch} and we use the public implementation and pre-trained model checkpoints of DINO~\cite{dino}.~\footnote{\url{https://github.com/facebookresearch/dino}.}
All the experiments with DINO ViTs use DINO ViT-small with a patch size of 8$\times$8.
Input images to the model are resized to $400 \times 400$ without any data augmentation schemes, and are fed to a ViT-small of 8$\times$8 patch size, which returns $50^2$ patch tokens.
We resize the support image token map dimension, originally $H_\texts \times W_\texts$, to 12$\times$12 via bilinear interpolation to reduce memory footprint.
The architecture details of CST are illustrated in \figref{fig:corrtrlayer} and enumerated in \tableref{table:corrtrlayer}.
Pascal-$5^{i}$ and COCO-$20^{i}$ are derived from Pascal Visual Object Classes 2012~\cite{pascal} and Microsoft Common Object in Context 2014~\cite{coco}, respectively.
All experiments use four NVIDIA Tesla V100 GPUs.

\subsection{Further analyses}
In this section we provide supplementary analyses and results omitted in the main paper due to the page limit.

\lessmarginparagraph{Computational complexity.}
Table~\ref{table:complexity} compares the numbers of parameters, MACs, GPU memory consumption of different methods.
The computational complexity of each model is evaluated on forwarding a 1-way 1-shot episode of images with $400\times400$ size.

\begin{figure}[t!]
	\centering
	\small
    \includegraphics[width=\linewidth]{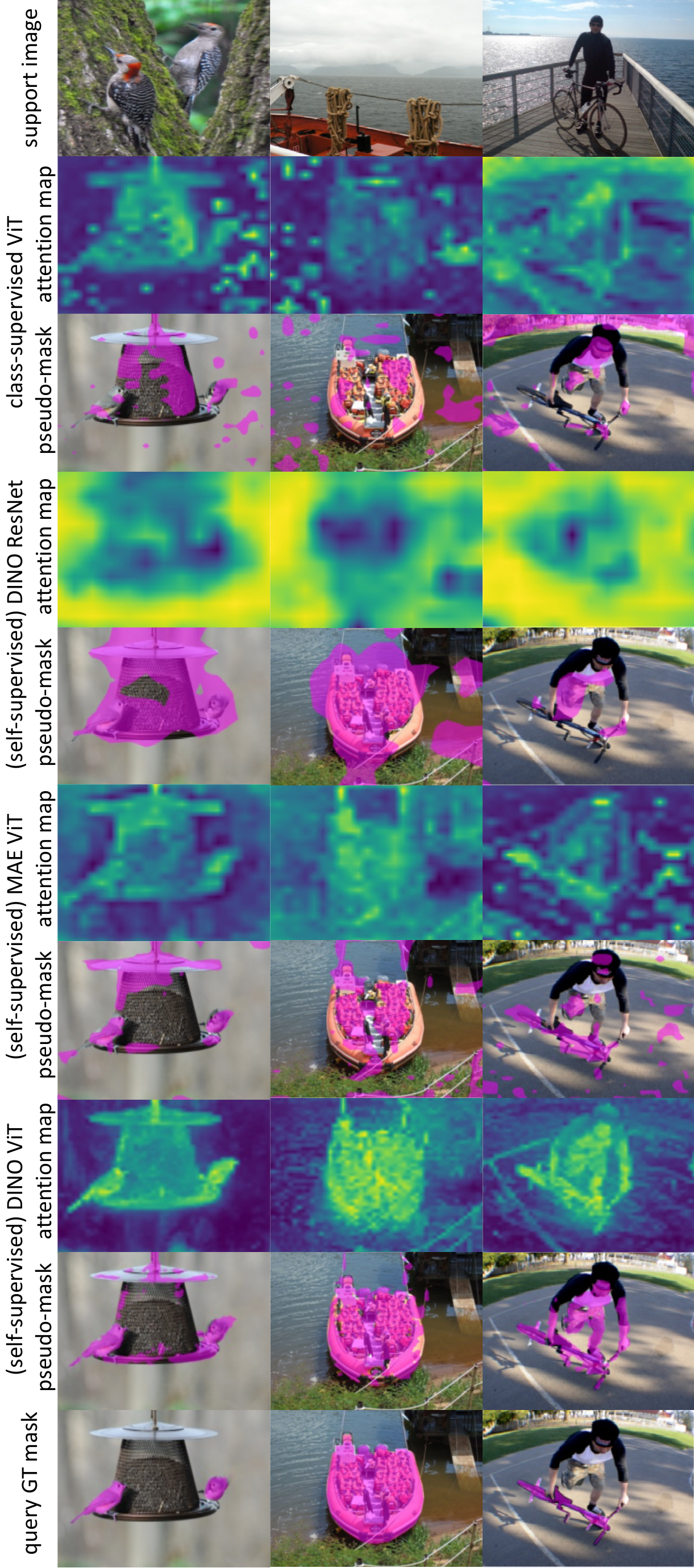}
	\caption{Pseudo-GT masks generated from class-supervised ViT~\cite{vit}, self-supervised DINO ResNet~\cite{resnet, dino}, self-supervised MAE ViT~\cite{mae, vit}, and self-supervised DINO ViT~\cite{dino, vit} from the top.
    }
\label{fig:different_bb_pmasks}
\vspace{-3mm}
\end{figure}

\lessmarginparagraph{Effect of self- vs. class-supervised ViT backbone.}
Table~\ref{table:self_class_supervised_dino} shows the superior performance of the self-supervised ViT backbone when compared with the class-supervised one, on the task of FS-CS.
The class-supervised ViT has the same architecture as the self-supervised one, but was trained on the 1000-class classification task using class supervision on ImageNet 1K~\cite{russakovsky2015imagenet}.
The gap is especially significant when training our CST model using only image-level labels (33.2\% vs. 6.9\%).
The class-supervised ViTs localize foreground regions less accurately than the self-supervised one (\cf \figref{fig:different_bb_pmasks}), leading to less accurate pseudo-labels and, ultimately, lower segmentation performance.

\begin{table}[!t]
    \small
    \centering
    \centering
    \scalebox{0.88}{
        \begin{tabular}{lclcr}
        \toprule
        name & $C_{\text{in}}\rightarrow C_{\text{out}}$ & component & count & \# params \\ \midrule
        correlation & \multirow{2}{*}{72$\rightarrow$128} & correlation & \multirow{2}{*}{2} & \multirow{2}{*}{77.5 K} \\ 
        transformer & & transformer layers & & \\ \hline
        clf. head & 128$\rightarrow$2 & 1x1 convolutions & 2 & 29.1 K \\ \hline
        seg. head & 128$\rightarrow$2 & 3x3 convolutions & 2 & 259.5 K \\
        \midrule
        total & & & & 366.0 K \\
         \bottomrule
        \end{tabular}
    }
    \vspace{-2mm}
    \caption{Model components of Classification-Segmentation Transformer (\ournet). Input and output channel dimensions are denoted with $C_{\text{in}}$ and $C_{\text{out}}$. 
    The backbone network is omitted.
    }
    \label{table:corrtrlayer}
    \vspace{-5mm}
\end{table}

\begin{table}[!t]
    \small
    \centering
    \setlength{\tabcolsep}{4pt}
    \scalebox{0.85}{
        \begin{tabular}{lcccccc}
        \toprule
        \multirow{2}{*}{model} & \multicolumn{3}{c}{frozen backbone} & \multicolumn{2}{c}{training module} & GPU \\ 
        
        \cmidrule(lr){2-4}\cmidrule(lr){5-6}
        & name & MACs & \# params & MACs & \# params & memory \\ \midrule
        HSNet~\cite{hsnet}  & ResNet50 & 13.4 G & 23.6 M & 17.7 G & 2.6 M & 2.1 G \\
        ASNet~\cite{ifsl}   & ResNet50 & 13.4 G & 23.6 M & 7.3 G & 1.3 M & 2.3 G \\
        CST (ours)          & ViT-S/8 & 53.4 G & 21.7 M & 3.7 G & 0.4 M & 2.4 G \\
         \bottomrule
        \end{tabular}
    }
    \vspace{-2mm}
    \caption{Comparing computational complexity of different models for forwarding a 1-way 1-shot episode.
    }
    \label{table:complexity}
    \vspace{-3mm}
\end{table}

\begin{table}[t!]
    \centering
    \small
    \scalebox{0.98}{
        \begin{tabular}{llcc}
            \toprule
             ViT backbone & superv. & clf. ER & seg. mIoU \\
             \midrule
             class-sup. & image & 79.6 & 6.9  \\
             \ccol self-sup. (\ournet) & \ccol image & \ccol \textbf{79.9} & \ccol \textbf{33.2} \\
             class-sup. & pixel & 85.8 & 54.0  \\
             \ccol self-sup. (\ournet) & \ccol pixel & \ccol \textbf{85.7} & \ccol \textbf{55.5} \\
             \bottomrule
        \end{tabular}
    }
    \vspace{-0mm}
    \caption{
    Comparing class- and self-supervised ViT backbones for FS-CS on Pascal-5$^{i}$~\cite{shaban2017oslsm}.
    }
    \label{table:self_class_supervised_dino}
    \vspace{-2mm}
\end{table}

\begin{table}[t!]
    \centering
    \small
    \scalebox{0.95}{
        \begin{tabular}{lllccr}
            \toprule
             & & & clf. & seg. & learn. \\
             method & backbone & task heads & ER & mIoU & params. \\
             \midrule
             ASNet~\cite{ifsl} & ResNet50 & coupled & 84.9 & 52.3 & 1.4M \\
             \ournet-(a) & ResNet50 & coupled & 83.9 & 51.0 & 0.4M \\
             \ournet-(b) & ResNet50 & decoupled & 84.1 & 50.9 & 0.4M  \\
             \ournet-(c) & DINO ViT & coupled & 84.3 & 54.2 & 0.4M \\
             \ccol \ournet & \ccol DINO ViT & \ccol decoupled & \ccol \textbf{85.7} & \ccol \textbf{55.5} & \ccol 0.4M \\
             \bottomrule
        \end{tabular}
    }
    \vspace{-0mm}
    \caption{
    Comparing self-supervised DINO ViT and class-supervised ResNet50 for FS-CS with image-level supervision on Pascal-5$^{i}$.
    }
    \label{table:vitsmall_vs_resnet50}
    \vspace{-0mm}
\end{table}

\begin{table}[!t]
	\small
	\centering
	\tabcolsep=0.1cm
	{
    \scalebox{0.88}{
		\begin{tabular}{ 
				l | c c | c c
				 }
			\toprule
             & \multicolumn{2}{c}{image-level}  & \multicolumn{2}{c}{pixel-level}  \\
            model & clf. ER (\%) & seg. mIoU (\%) & clf. ER (\%) & seg. mIoU (\%) \\ \hline
            HSNet*~\cite{hsnet}  & 76.4 & 31.7 & 82.2 & 49.5 \\
			ASNet*~\cite{ifsl}   & 78.6 & 32.8 & 84.7 & 53.7 \\
            CST (ours)          & \textbf{79.9} & \textbf{33.2} & \textbf{85.7} & \textbf{55.5} \\
			\bottomrule      			                             
		\end{tabular}
    }
	}
	\vspace{-2mm}
	\caption{
 Comparing model performance on FS-CS using the DINO ViT backbone on Pascal-5$^{i}$. 
 The methods denoted with * are reproduced with the same DINO ViT backbone that CST uses.
 In image-level supervised learning, all methods are trained with the generated pseudo-GT mask labels.}
	\label{table:pdinovit}   	
\end{table}

\lessmarginparagraph{Comparing DINO ViT-small vs. ResNet50.}
Table~\ref{table:vitsmall_vs_resnet50} compares results using ResNet50~\cite{resnet} or ViT-small as backbones.
Note that \ournet has independent classification and segmentation task heads that take input from class and image tokens respectively. 
Therefore, it is not trivial to use ResNet50 as a backbone for \ournet in order to compare to ResNet-based methods~\cite{tian2020pfenet, hsnet, ifsl}.
Albeit not an apple-to-apple comparison, we adapt ResNet50 for \ournet by artificially creating a ``class token'' for representing global image semantics, by global average-pooling ResNet feature maps.
The ResNet-based CSTs (\ournet-(a) and (b)) achieve similar performance to ASNet~\cite{ifsl} with fewer learnable parameters.
Comparing \ournet-(a) and (b), the gain from using decoupled task heads is unclear, unlike with the ViT-based \ournet. 
We hypothesize that, because the class and segmentation representations are generated from the same ResNet features, there is less benefit of task-head specialization when compared to using the class and image tokens in ViTs.

Similarly, Table~\ref{table:pdinovit} compares results using DINO ViT-small as the backbones.
Other methods~\cite{hsnet, ifsl} are reproduced with the DINO ViT backbone such that they take the same correlation tokens with CST.
As those methods~\cite{hsnet, ifsl} benefit from fusing the pyramidal ResNet features, they are not perfectly compatible with DINO ViTs.

\lessmarginparagraph{Comparison with other self-supervised backbones.}
Table~\ref{table:ssbackbone} compares the performance of our method when using different self-supervised backbones on FS-CS:
DINO ResNet50 and Masked Auto-encoder (MAE) ViT~\cite{mae}.
MAE learns to reconstruct some masked input image tokens in an auto-encoder~\cite{autoencoderorigin, denoisingautoencoder} framework, requiring no supervision.
We use an MAE-trained model with an architecture that is identical to the DINO-trained ViTs, and that is also publicly available~\footnote{\url{https://github.com/facebookresearch/mae}. As the available models are pre-trained with 16$\times$16 patches, we compare it with DINO ViTs pretrained with the same patch size.}.
We observe that MAE ViTs show weaker localization properties when compared to DINO ViTs as qualitatively compared in \figref{fig:different_bb_pmasks}, resulting in low segmentation performance.

\begin{table}[t!]
    \centering
    \small
    \scalebox{0.98}{
        \begin{tabular}{llcc}
            \toprule
             backbone & pretrained with & clf. ER & seg. mIoU \\
             \midrule
             ViT~\cite{vit} & MAE~\cite{mae} & 68.5 & 13.4  \\
             ViT~\cite{vit} & DINO~\cite{dino} & 82.2 & 33.7  \\
             \bottomrule
        \end{tabular}
    }
    \vspace{-0mm}
    \caption{
    Comparing self-supervised DINO and MAE backbones for FS-CS with image-level supervision on Pascal-5$^{i}$.
    }
    \label{table:ssbackbone}
    \vspace{-3mm}
\end{table}

\lessmarginparagraph{Effect of multi-head token correlations.}
Table~\ref{table:multihead_tokencorrelations} shows the effectiveness of head-wise correlation tokens in \eqref{eq:headwisecorr}. Table~\ref{table:multihead_tokencorrelations}.
We split the class and image tokens into $M$ equal-sized tokens, each with dimensionality $C/M$, compute cross-correlation, and then concatenate the $M$-head correlations. 
We observe that the multi-head token correlation is not only effective in performance but also boosting faster convergence.

\lessmarginparagraph{Loss-balancing hyperparameter $\lambda$.}
We choose the loss-balancing hyperparameter $\lambda$ based on the course-grained hyperparameter search shown in Table~\ref{table:loss_hyperparam}.
The experimental results show that the performance does not differ significantly, implying that it is robust to different values of $\lambda$.

\lessmarginparagraph{Four-fold performance.}
We omit the four-fold performance for a few experiments in the main paper due to space limitations, and thus specify the four-fold performance of Tables~\ref{table:c_cls}, \ref{table:c_mask}, \ref{table:c_fss}, and \figref{fig:p-nway} of the main paper in Tables~\ref{table:multiway}, \ref{table:c_cls_foldwise}, \ref{table:c_mask_foldwise}, \ref{table:c_fss_foldwise}, respectively, for reference.

\lessmarginparagraph{Experiments with $K>1$ shots.}
Tables~\ref{table:p_kshot} and \ref{table:c_kshot} present FS-CS performance when 5-shot support examples are used for each class during testing. 
Using five shots per class brings 3.0\% segmentation improvement compared to one-shot models, when zero ground-truth pixel-level labels are used.
It is noteworthy that using more shots leads to greater performance gains when learning with pseudo-GT masks, when compared to learning without it;
\ournet with pseudo-GT masks (\checkmark) improves from 33.2\% $\rightarrow$ 36.2\% with 1 $\rightarrow$ 5 shots, \ournet without pseudo-GT masks (\xmark) improves from 16.0\% $\rightarrow$ 16.9\% (\cf \tableref{table:p_cls} of the main paper).

\begin{figure}[t!]
	\centering
	\small
    \includegraphics[width=\linewidth]{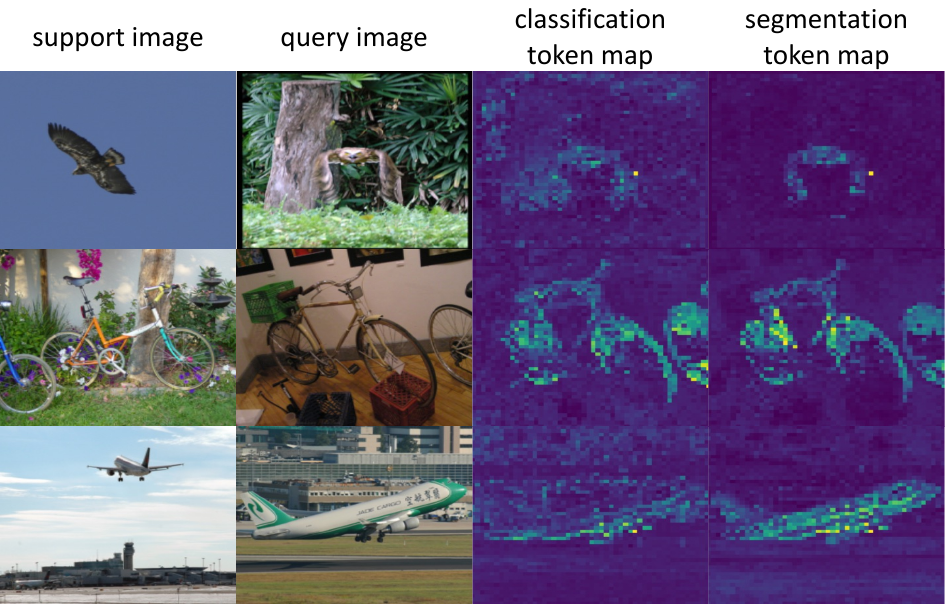}
	\caption{Examples of channel-averaged classification and segmentation task token maps, $\bC_\textclf$ and $\bC_\textseg$.
    }
\label{fig:clfseg}
\vspace{-3mm}
\end{figure}

\lessmarginparagraph{Visualization of classification and segmentation token maps.}
Figure~\ref{fig:clfseg} visualizes more samples of the channel-averaged classification and segmentation token maps, $\bC_\textclf$ and $\bC_\textseg$, some of which were also visualized in \figref{fig:twoheads} of the main paper.
The segmentation token maps delineate object boundaries and the foreground, while suppressing background regions.

\lessmarginparagraph{Visualization of pseudo-GT masks generated from different pre-trained feature extractors.}
Figure~\ref{fig:different_bb_pmasks} visualizes the pseudo-GT masks generated from different models.
Pseudo-GT masks from three ViTs with different training procedures are produced as formulated in equations \ref{eq:pmask}-\ref{eq:pmask_thr} of the main paper.
To generate pseudo-labels from a self-supervised ResNet, the globally-average pooled feature map from the support image is correlated against an image feature map.
In addition, we qualitatively observe that, with the ResNet backbone, the inverse correlation better captures foreground objects, and visualize this instead.
The qualitative results align with the quantitative results (Table 6 in the main paper), in that the DINO ViT backbone produces the highest-quality pseudo-GT masks.

\begin{figure}[t!]
	\centering
	\small
    \includegraphics[width=0.95\linewidth]{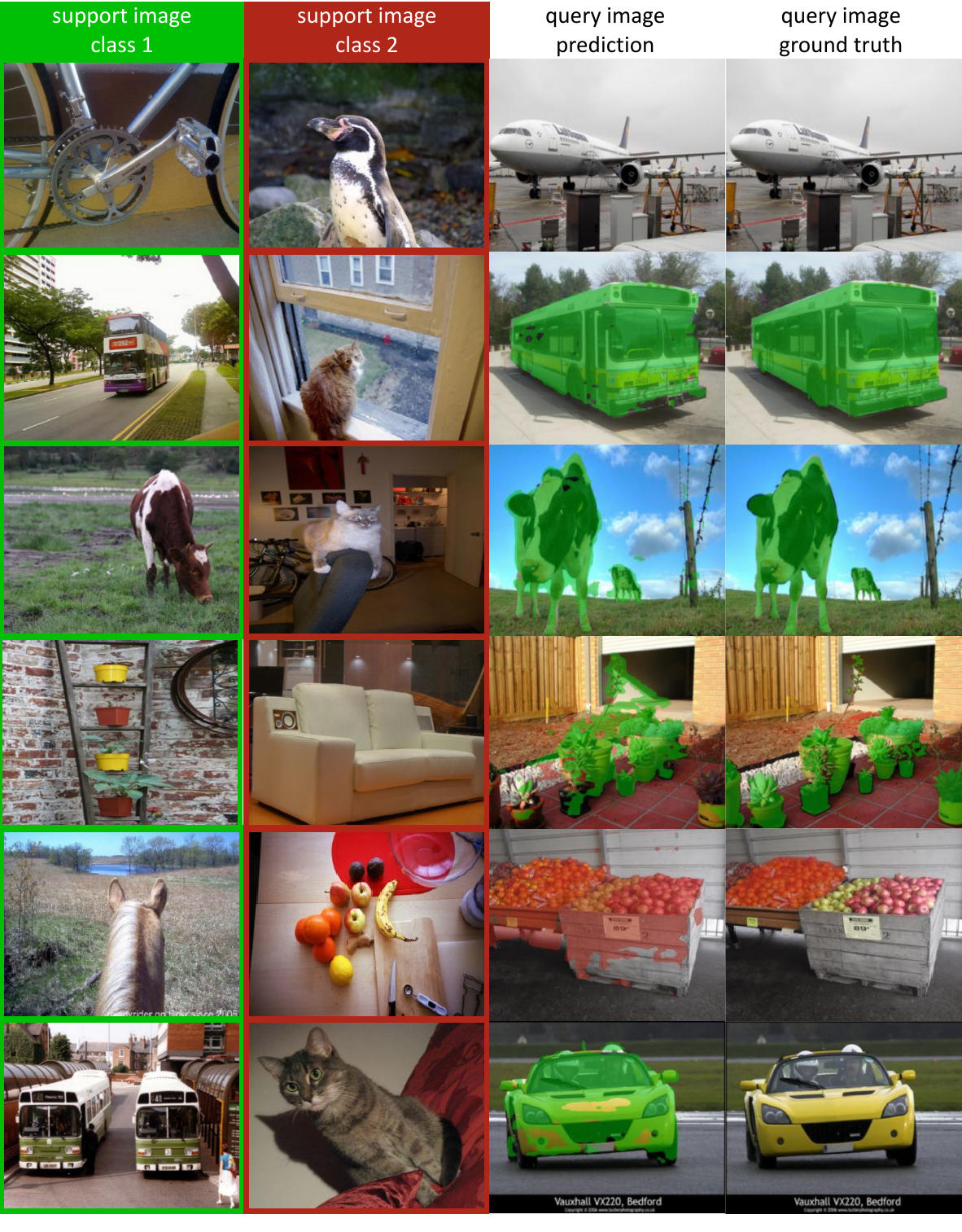}
    \vspace{-3mm}
	\caption{2-way 1-shot segmentation prediction of \ournet trained with image-level labels. Image frames on the support images distinguish classes by colors.
    }
\label{fig:weaksup_final}
\end{figure}

\begin{figure}[t!]
	\centering
	\small
    \includegraphics[width=0.95\linewidth]{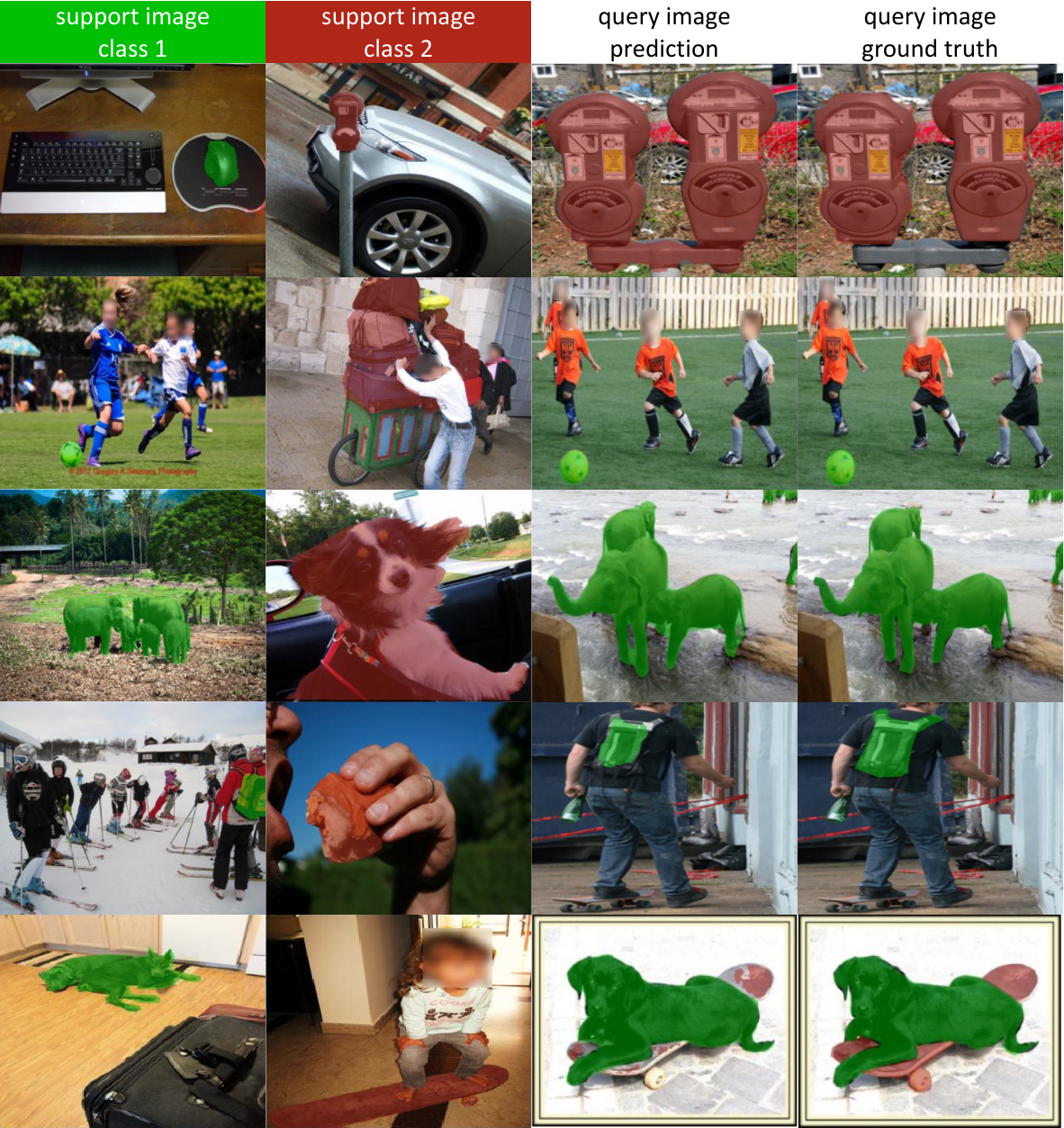}
	\caption{2-way 1-shot segmentation prediction of \ournet. The model is trained with pixel-level labels during training, and pixel-level support annotations are also given during testing (as overlayed on the support images with colors). 
 Human faces are anonymized for visualization. 
    }
\label{fig:masksup_final}
\end{figure}

\lessmarginparagraph{Visualization of predicted segmentation masks.}
Figure~\ref{fig:weaksup_final} visualizes segmentation masks predicted by the image-level supervised \ournet.
The model correctly segments no foreground when support image classes are irrelevant to those of the query (1st row).
Note that the model can also segment small objects, and multiple objects.
The last two rows show failure cases;
the model segments the apple container which is not present in the ground-truth annotation (5th row)
and incorrectly segments a car given the support image containing buses, which may result from vehicular class confusion (6th row).
Figure~\ref{fig:masksup_final} visualizes segmentation masks predicted by the pixel-level supervised \ournet.
Leveraging pixel-level ground-truth annotations, the pixel-level supervised model captures small objects at image corners (2nd and 3rd row) and multiple objects from multiple classes (5th row) precisely.

\begin{table}[t!]
    \centering
    \small
    \scalebox{0.98}{
        \begin{tabular}{llcc}
            \toprule
             correlation  & superv. & clf. ER & seg. mIoU \\
             \midrule
             single-head & image & \textbf{80.3} & 33.1  \\
             \ccol multi-head (\ournet) & \ccol image & \ccol 79.9 & \ccol \textbf{33.2} \\
             single-head & pixel & 85.5 & 54.3 \\
             \ccol multi-head (\ournet) & \ccol pixel & \ccol \textbf{85.7} & \ccol \textbf{55.5} \\
             \bottomrule
        \end{tabular}
    }
    \vspace{-0mm}
    \caption{
    Comparing single- and multi-head token correlations of \eqref{eq:headwisecorr} for FS-CS on Pascal-5$^{i}$.
    }
    \label{table:multihead_tokencorrelations}
    \vspace{-3mm}
\end{table}

\begin{table}[t!]
    \centering
    \small
    \setlength{\tabcolsep}{5pt}\
    \scalebox{0.95}{
        \begin{tabular}{lccc}
            \toprule
             hyperparameter $\lambda$ & 0.05 & 0.1 & 0.5 \\ \midrule
             clf. ER   (\%) & 79.6 & 79.9 & 80.5 \\
             seg. mIoU (\%) & 33.1 & 33.2 & 32.9\\
             \bottomrule
        \end{tabular}
    }
    \vspace{-0mm}
    \caption{
    Hyperparameter search on the loss balancing hyperparameter $\lambda$ in \eqref{eq:loss_final} in the main paper.
    }
    \label{table:loss_hyperparam}
    \vspace{-0mm}
\end{table}

\begin{table}[t!]
    \centering
    \small
    \setlength{\tabcolsep}{4.3pt}
    \scalebox{0.78}{
        \begin{tabular}{lccccc||ccccc}
            \toprule
             & \multicolumn{10}{c}{$N$-way 1-shot}  \\
             \cmidrule(lr){2-11} 
             & \multicolumn{5}{c}{classification 0/1 exact ratio (\%)} & \multicolumn{5}{c}{segmentation mIoU (\%)} \\
             \cmidrule(lr){2-6}\cmidrule(lr){7-11} 
             method & 1 & 2 & 3 & 4 & 5 & 1 & 2 & 3 & 4 & 5 \\
             \midrule
             PANet \cite{wang2019panet}     & 69.0 & 50.9 & 39.3 & 29.1 & 22.2 & 36.2 & 37.2 & 37.1 & 36.6 & 35.3 \\
             PFENet \cite{tian2020pfenet}   & 74.6 & 41.0 & 24.9 & 14.5 &  7.9 & 43.0 & 35.3 & 30.8 & 27.6 & 24.9 \\
             HSNet \cite{hsnet}             & 82.7 & 67.3 & 52.5 & 45.2 & 36.8 & 49.7 & 43.5 & 39.8 & 38.1 & 36.2 \\
             ASNet \cite{ifsl}              & 84.9 & 68.3 & 55.8 & 46.8 & \textbf{37.3} & 52.3 & 47.8 & 45.4 & 44.5 & 42.4 \\
             \ournet                        & \textbf{85.7} & \textbf{70.4} & \textbf{57.3} & \textbf{47.3} & 36.9 & \textbf{55.5} & \textbf{53.7} & \textbf{52.6} & \textbf{52.0} & \textbf{50.3} \\
             \bottomrule
        \end{tabular}
    }
    \vspace{-2mm}
    \caption{
    Numerical performance of \figref{fig:p-nway} in the main paper:
    performance comparison on FS-CS with pixel-level supervision on $N$-way 1-shot Pascal-5$^i$.
}
    \label{table:multiway}
    \vspace{-2mm}
\end{table}

\begin{table*}[t!]
    \centering
    \small
    \setlength{\tabcolsep}{5pt}
    \scalebox{0.75}{
        \centering
        \begin{tabular}{lcccccc|ccccc||ccccc|ccccc|r}
            \toprule
             & &\multicolumn{10}{c}{1-way 5-shot} & \multicolumn{10}{c}{2-way 5-shot}\\
             \cmidrule(lr){3-12}\cmidrule(lr){13-22}
             & & \multicolumn{5}{c}{classification 0/1 exact ratio (\%)} & \multicolumn{5}{c}{segmentation mIoU (\%)} & \multicolumn{5}{c}{classification 0/1 exact ratio (\%)} & \multicolumn{5}{c}{segmentation mIoU (\%)} & learn. \\
             \cmidrule(lr){3-7}\cmidrule(lr){8-12}\cmidrule(lr){13-17}\cmidrule(lr){18-22}  method & ps-mask & $5^{0}$ & $5^{1}$ & $5^{2}$ & $5^{3}$ & avg. & $5^{0}$ & $5^{1}$ & $5^{2}$ & $5^{3}$ & avg. & $5^{0}$ & $5^{1}$ & $5^{2}$ & $5^{3}$ & avg. & $5^{0}$ & $5^{1}$ & $5^{2}$ & $5^{3}$ & avg. & params. \\
             \midrule
             HSNet~\cite{hsnet} & \xmark & 89.3 & \textbf{90.1} & 66.3 & \textbf{90.7} & 84.1 & 12.5 & 24.7 & 19.4 & 18.1 & 18.7 & 81.3 & 78.4 & 44.0 & \textbf{81.4} & 71.3 & 13.0 & 25.4 & 22.2 & 18.7 & 19.8 & 2.6M \\
             ASNet~\cite{ifsl}  & \xmark & 84.3 & 89.1 & 66.2 & 90.0 & 82.4 & 11.5 & 22.0 & 14.0 & 17.4 & 16.2 & 72.5 & \textbf{80.6} & 41.8 & 76.8 & 67.9 & 8.7 & 23.1 & 11.8 & 18.0 & 15.4 & 1.3M  \\
             \ournet            & \xmark & 88.8 & 85.1 & 63.8 & 88.7 & 81.6 & 13.1 & 21.6 & 15.3 & 17.6 & 16.9 & \textbf{88.8} & 74.2 & 41.6 & 78.9 & 70.9 & 13.1 & 22.3 & 15.6 & 17.5 & 17.1 & 0.4M \\
             DINO~\cite{dino} & $\diamond$ & - & - & - & - & - & 20.1 & 23.6 & 16.4 & 16.8 & 19.2 & - & - & - & - & - & 12.9 & 11.9 & 8.4 & 9.4 & 10.7 & 0 \\
             \ournet & \checkmark        & \textbf{92.7} & 89.4 & \textbf{70.3} & 89.2 & \textbf{85.4} & \textbf{42.1} & \textbf{40.8} & \textbf{30.8} & \textbf{31.2} & \textbf{36.2} & 86.2 & 77.4 & \textbf{48.5} & 73.9 & \textbf{71.5} & \textbf{40.9} & \textbf{40.1} & \textbf{29.8} & \textbf{31.3} & \textbf{35.5} & 0.4M \\
             \bottomrule
        \end{tabular}
    }
    \vspace{-2mm}
    \caption{Comparing 5-shot performance, \ie, 5 support images per class, on FS-CS trained with image-level supervision on Pascal-5$^i$. 
}
    \label{table:p_kshot}
\end{table*}

\begin{table*}[t!]
    \centering
    \small
    \setlength{\tabcolsep}{5pt}
    \scalebox{0.75}{
        \centering
        \begin{tabular}{lcccccc|ccccc||ccccc|ccccc|r}
            \toprule
             & &\multicolumn{10}{c}{1-way 5-shot} & \multicolumn{10}{c}{2-way 5-shot}\\
             \cmidrule(lr){3-12}\cmidrule(lr){13-22}
             & & \multicolumn{5}{c}{classification 0/1 exact ratio (\%)} & \multicolumn{5}{c}{segmentation mIoU (\%)} & \multicolumn{5}{c}{classification 0/1 exact ratio (\%)} & \multicolumn{5}{c}{segmentation mIoU (\%)} & learn. \\
             \cmidrule(lr){3-7}\cmidrule(lr){8-12}\cmidrule(lr){13-17}\cmidrule(lr){18-22}  method & ps-mask & $5^{0}$ & $5^{1}$ & $5^{2}$ & $5^{3}$ & avg. & $5^{0}$ & $5^{1}$ & $5^{2}$ & $5^{3}$ & avg. & $5^{0}$ & $5^{1}$ & $5^{2}$ & $5^{3}$ & avg. & $5^{0}$ & $5^{1}$ & $5^{2}$ & $5^{3}$ & avg. & params. \\
             \midrule
             \ournet            & \xmark & \textbf{79.3} & \textbf{83.4} & 83.2 & \textbf{86.3} & 83.1 & 11.9 & 11.2 & 8.4 & 11.1 & 10.6 & \textbf{64.9} & \textbf{73.6} & 72.3 & 72.6 & 70.9 & 10.7 & 11.2 & 8.3 & 10.7 & 10.2 & 0.4M \\
             DINO~\cite{dino} & $\diamond$ & - & - & - & - & - & 13.9 & 12.3 & 10.4 & 11.9 & 12.1 & - & - & - & - & - & 7.9 & 7.0 & 6.6 & 7.0 & 7.1 & 0 \\
             \ournet & \checkmark        & 78.8 & 83.3 & \textbf{86.7} & 84.9 & \textbf{83.4} & \textbf{20.8} & \textbf{20.9} & \textbf{20.6} & \textbf{21.1} & \textbf{20.8} & 64.3 & 71.7 & \textbf{77.3} & \textbf{72.8} & \textbf{71.5} & \textbf{19.0} & \textbf{21.0} & \textbf{20.8} & \textbf{20.9} & \textbf{20.4} & 0.4M \\
             \bottomrule
        \end{tabular}
    }
    \vspace{-2mm}
    \caption{Comparing 5-shot performance, \ie, 5 support images per class, on FS-CS trained with image-level supervision on COCO-20$^i$. 
}
    \label{table:c_kshot}
\end{table*}

\begin{table*}[t!]
    \centering
    \small
    \setlength{\tabcolsep}{5pt}
    \scalebox{0.75}{
        \centering
        \begin{tabular}{lcccccc|ccccc||ccccc|ccccc|r}
            \toprule
             & &\multicolumn{10}{c}{1-way 1-shot} & \multicolumn{10}{c}{2-way 1-shot}\\
             \cmidrule(lr){3-12}\cmidrule(lr){13-22}
             & & \multicolumn{5}{c}{classification 0/1 exact ratio (\%)} & \multicolumn{5}{c}{segmentation mIoU (\%)} & \multicolumn{5}{c}{classification 0/1 exact ratio (\%)} & \multicolumn{5}{c}{segmentation mIoU (\%)} & learn. \\
             \cmidrule(lr){3-7}\cmidrule(lr){8-12}\cmidrule(lr){13-17}\cmidrule(lr){18-22}  method & ps-mask & $5^{0}$ & $5^{1}$ & $5^{2}$ & $5^{3}$ & avg. & $5^{0}$ & $5^{1}$ & $5^{2}$ & $5^{3}$ & avg. & $5^{0}$ & $5^{1}$ & $5^{2}$ & $5^{3}$ & avg. & $5^{0}$ & $5^{1}$ & $5^{2}$ & $5^{3}$ & avg. & params. \\
             \midrule
             \ournet & \xmark & \textbf{74.2} & \textbf{78.4} & 65.9 & \textbf{79.6} & 74.5 & 11.8 & 10.7 & 8.1 & 10.6 & 10.3 & \textbf{60.4} & 60.2 & 67.5 & \textbf{61.2} & 62.3 & 10.3 & 10.4 & 7.8 & 9.4 & 9.5 & 0.4M \\
             DINO~\cite{dino} & $\diamond$ & - & - & - & - & - & 13.9 & 12.2 & 10.4 & 11.9 & 12.1 & - & - & - & - & - & 7.8 & 7.4 & 6.8 & 7.5 & 7.4 & 0 \\
             \ournet & \checkmark & 74.0 & \textbf{78.4} & \textbf{82.1} & 78.1 & \textbf{78.2} & \textbf{20.2} & \textbf{19.8} & \textbf{19.1} & \textbf{19.5} & \textbf{19.6} & 59.8 & \textbf{60.5} & \textbf{68.2} & \textbf{61.2} & \textbf{62.4} & \textbf{19.0} & \textbf{17.5} & \textbf{18.1} & \textbf{18.5} & \textbf{18.3} & 0.4M \\
             \bottomrule
        \end{tabular}
    }
    \vspace{-2mm}
    \caption{Four-fold results on FS-CS with image-level supervision on COCO-20$^i$. The results correspond to \tableref{table:c_cls} in the main paper.
}
    \label{table:c_cls_foldwise}
\end{table*}

\begin{table*}[t!]
    \centering
    \small
    \setlength{\tabcolsep}{5pt}
    \scalebox{0.86}{
        \begin{tabular}{lccccc|ccccc||ccccc|cccccr}
            \toprule
             & \multicolumn{10}{c}{1-way 1-shot} & \multicolumn{10}{c}{2-way 1-shot} \\
             \cmidrule(lr){2-11}\cmidrule(lr){12-21}
             & \multicolumn{5}{c}{classification 0/1 exact ratio (\%)} & \multicolumn{5}{c}{segmentation mIoU (\%)} & \multicolumn{5}{c}{classification 0/1 exact ratio (\%)} & \multicolumn{5}{c}{segmentation mIoU (\%)} \\
             \cmidrule(lr){2-6}\cmidrule(lr){7-11}\cmidrule(lr){12-16}\cmidrule(lr){17-21}
             method & $20^{0}$ & $20^{1}$ & $20^{2}$ & $20^{3}$ & avg. & $20^{0}$ & $20^{1}$ & $20^{2}$ & $20^{3}$ & avg. & $20^{0}$ & $20^{1}$ & $20^{2}$ & $20^{3}$ & avg. & $20^{0}$ & $20^{1}$ & $20^{2}$ & $20^{3}$ & avg. \\
             \midrule
             PANet \cite{wang2019panet}     & 64.3 & 66.5 & 68.0 & 67.9 & 66.7 & 25.5 & 24.7 & 25.7 & 24.7 & 25.2 & 42.5 & 49.9 & 53.6 & 47.8 & 48.5 & 24.9 & 25.0 & 23.3 & 21.4 & 23.6 \\
             PFENet \cite{tian2020pfenet}   & 70.7 & 70.6 & 71.2 & 72.9 & 71.4 & 30.6 & 34.8 & 29.4 & 32.6 & 31.9 & 35.6 & 34.3 & 43.1 & 32.8 & 36.5 & 23.3 & 23.8 & 20.2 & 23.1 & 22.6 \\
             HSNet \cite{hsnet}             & 74.7 & 77.2 & 78.5 & 77.6 & 77.0 & 36.2 & 34.3 & 32.9 & 34.0 & 34.3 & 57.7 & 62.4 & 67.1 & \textbf{62.6} & 62.5 & 28.9 & 29.6 & 30.3 & 29.3 & 29.5 \\
             ASNet \cite{ifsl} & 76.2 & 78.8 & 79.2 & 80.2 & 78.6 & 35.7 & 36.8 & 35.3 & 35.6 & 35.8 & 59.5 & 61.5 & \textbf{68.8} & 62.4 & 63.1 & 29.8 & 33.0 & 33.4 & 30.4 & 31.6 \\
             \ournet           & \textbf{77.6} & \textbf{82.0} & \textbf{83.1} & \textbf{80.5} & \textbf{80.8} & \textbf{36.3} & \textbf{38.3} & \textbf{37.8} & \textbf{40.7} & \textbf{38.3} & \textbf{61.0} & \textbf{66.0} & 68.2 & 60.5 & \textbf{64.0} & \textbf{34.7} & \textbf{37.1} & \textbf{36.8} & \textbf{36.3} & \textbf{36.2} \\

             \bottomrule
        \end{tabular}
    }
    \vspace{-2mm}
    \caption{
    Four-fold results on FS-CS with pixel-level supervision on COCO-20$^i$.
    The results correspond to \tableref{table:c_mask} in the main paper.
}
    \label{table:c_mask_foldwise}
\end{table*}

\begin{table*}[t!]
    \centering
    \small
    \scalebox{0.97}{
        \begin{tabular}{lccccccccccccr}
            \toprule
             & \multicolumn{6}{c}{1-way 1-shot} & \multicolumn{6}{c}{1-way 5-shot} & \small{\# learn.} \\
             \cmidrule(lr){2-7}\cmidrule(lr){8-13}
             method & $20^{0}$ & $20^{1}$ & $20^{2}$ & $20^{3}$ & mIoU & FBIoU & $20^{0}$ & $20^{1}$ & $20^{2}$ & $20^{3}$ & mIoU & FBIoU & \small{params.}\\
             \midrule
             RPMM \cite{yang2020pmm} & 29.5 & 36.8 & 28.9 & 27.0 & 30.6 & - & 33.8 & 42.0 & 33.0 & 33.3 & 35.5 & - & 38.6 M \\ 
             RePRI \cite{malik2021repri} & 31.2 & 38.1 & 33.3 & 33.0 & 34.0 & - & 38.5 & 46.2 & 40.0 & 43.6 & 42.1 & - & - \\ 
             SSP \cite{fan2022self} & 35.5 & 39.6 & 37.9 & 36.7 & 37.4 & - & 40.6 & 47.0 & 45.1 & 43.9 & 44.1 & - & 8.7M \\
             MMNet \cite{wu2021learning} & 34.9 & 41.0 & 37.2 & 37.0 & 37.5 & - & 37.0 & 40.3 & 39.3 & 36.0 & 38.2 & - & 10.4 M \\ 
             MLC \cite{yang2021mining} & \textbf{46.8} & 35.3 & 26.2 & 27.1 & 33.9 & - & \textbf{54.1} & 41.2 & 34.1 & 33.1 & 40.6 & - & 8.7 M \\ 
             NTRENet \cite{liu2022learning} & 36.8 & 42.6 & 39.9 & 37.9 & 39.3 & 68.5 & 38.2 & 44.1 & 40.4 & 38.4 & 40.3 & 69.2 & -\\
             CMN \cite{xie2021few} & 37.9 & 44.8 & 38.7 & 35.6 & 39.3 & 61.7 & 42.0 & 50.5 & 41.0 & 38.9 & 43.1 & 63.3 & - \\ 
             HSNet \cite{hsnet} & 36.3 & 43.1 & 38.7 & 38.7 & 39.2 & 68.2 & 43.3 & 51.3 & 48.2 & 45.0 & 46.9 & 70.7 & 2.6 M \\   
             DACM \cite{xiong2022doubly} & 37.5 & 44.3 & 40.6 & 40.1 & 40.6 & 68.9 & 44.6 & \textbf{52.0} & 49.2 & 46.4 & 48.1 & 71.6 & -\\
             ASNet \cite{ifsl} & 41.5 & 44.1 & 42.8 & 40.6 & 42.2 & 68.8 & 47.6 & 50.1 & 47.7 & 46.4 & 47.9 & 71.6 & 1.3 M \\
             \ournet & 39.6 & \textbf{45.8} & \textbf{45.0} & \textbf{45.5} & \textbf{44.0} & \textbf{70.3} & 42.8 & 51.6 & \textbf{50.2} & \textbf{50.2} & \textbf{48.7} & \textbf{73.7} & 0.4 M \\
             \bottomrule
        \end{tabular}
    }
    \vspace{-2mm}
    \caption{Four-fold performance on the conventional few-shot segmentation task (FS-S) with image-level supervision on COCO-20$^i$~\cite{nguyen2019fwb}. 
    The results correspond to \tableref{table:c_fss} in the main paper.}
    \label{table:c_fss_foldwise}
\end{table*}

\clearpage

\end{document}